\documentclass[10pt,twocolumn,letterpaper]{article}
\usepackage{graphicx}
\usepackage{amsmath}
\usepackage{amssymb}
\usepackage{booktabs}
\usepackage{enumitem}
\usepackage{verbatim}

\usepackage{subfigure}

\usepackage{wrapfig}
\usepackage{times}
\usepackage{epsfig}
\usepackage[utf8]{inputenc} 
\usepackage[T1]{fontenc}    
\usepackage[pagenumbers]{cvpr} 

\usepackage{siunitx}
\usepackage{cite}
\usepackage{ctable}
\usepackage{hhline}
\usepackage{booktabs}

\usepackage{csquotes}
\usepackage{pifont}
\usepackage{mathtools}
\usepackage{animate}

\usepackage{anyfontsize}
\usepackage[super]{nth}

\usepackage{ctable}
\usepackage{soul}
\usepackage{multicol}
\usepackage{makecell}
\usepackage[normalem]{ulem}
\usepackage{multirow}
\usepackage{float}
\usepackage{stfloats}

\usepackage{stmaryrd}
\usepackage{mathstyle}

\definecolor{Gray}{gray}{0.9}

\usepackage{xcolor}         
\usepackage{color, colortbl}
\definecolor{red}{RGB}{178,34,34}
\definecolor{blue}{RGB}{0,0,255}

\usepackage{xspace}
\usepackage{mmstyle}
\usepackage{multirow}
 \usepackage{threeparttable} 
\usepackage{subfigure} 
\usepackage[pagebackref,breaklinks,colorlinks]{hyperref}

\usepackage[capitalize]{cleveref}
\crefname{section}{Sec.}{Secs.}
\Crefname{section}{Section}{Sections}
\Crefname{table}{Table}{Tables}
\crefname{table}{Tab.}{Tabs.}

\newcommand{\modelname}{\emph{LTSF-Linear}\xspace}
\begin{document}

\title{Are Transformers Effective for Time Series Forecasting?}

\newcommand*{\affaddr}[1]{#1} 
\newcommand*{\affmark}[1][*]{\textsuperscript{#1}}
\newcommand*{\email}[1]{\texttt{#1}}

\makeatletter
\newcommand{\printfnsymbol}[1]{%
  \textsuperscript{\@fnsymbol{#1}}%
}
\makeatother

\author{%
Ailing Zeng\affmark[1]\thanks{Equal contribution} , Muxi Chen\affmark[1]\printfnsymbol{1}, Lei Zhang\affmark[2], Qiang Xu\affmark[1]\\
\affaddr{\affmark[1]The Chinese University of Hong Kong}\\
\affaddr{\affmark[2]International Digital Economy Academy (IDEA)}\\
\email{\tt\small\{alzeng, mxchen21, qxu\}@cse.cuhk.edu.hk}\\
\email{\tt\small\{leizhang\}@idea.edu.cn}
}


\maketitle

\begin{abstract}
Recently, there has been a surge of Transformer-based solutions for the long-term time series forecasting (LTSF) task. Despite the growing performance over the past few years, \emph{we question the validity of this line of research in this work}.
Specifically, Transformers is arguably the most successful solution to extract the semantic correlations among the elements in a long sequence. However, in time series modeling, we are to extract the temporal relations in \emph{an ordered set of continuous points}. While employing positional encoding and using tokens to embed sub-series in Transformers facilitate preserving some ordering information, the nature of the \emph{permutation-invariant} self-attention mechanism inevitably results in temporal information loss. 

To validate our claim, we introduce a set of embarrassingly simple one-layer linear models named \emph{LTSF-Linear} for comparison. Experimental results on nine real-life datasets show that \emph{LTSF-Linear} surprisingly outperforms existing sophisticated Transformer-based LTSF models in all cases, and often by a large margin. Moreover, we conduct comprehensive empirical studies to explore the impacts of various design elements of LTSF models on their temporal relation extraction capability.
We hope this surprising finding opens up new research directions for the LTSF task. We also advocate revisiting the validity of Transformer-based solutions for other time series analysis tasks (e.g., anomaly detection) in the future. Code is available at: \url{https://github.com/cure-lab/LTSF-Linear}.

\end{abstract}

\section{Introduction}
\label{sec:intro}

Time series are ubiquitous in today's data-driven world. Given historical data, time series forecasting (TSF) is a long-standing task that
has a wide range of applications, including but not limited to traffic flow estimation, energy management, and financial investment.
Over the past several decades, TSF solutions have undergone a progression from traditional statistical methods (e.g., ARIMA~\cite{ariyo2014arima}) and machine learning techniques (e.g., GBRT~\cite{gbrt}) to deep learning-based solutions, e.g., Recurrent Neural Networks~\cite{GuokunLai2017lstm} and Temporal Convolutional Networks~\cite{bai2018empirical,liu2021time}. 

Transformer~\cite{vaswani2017attention} is arguably the most successful sequence modeling architecture, demonstrating unparalleled performances in various applications, such as natural language processing (NLP)~\cite{devlin2018bert}, speech recognition~\cite{dong2018speech}, and computer vision~\cite{liu2021swin,zeng2022deciwatch}. 
Recently, there has also been a surge of Transformer-based solutions for time series analysis, as surveyed in ~\cite{wen2022transformers}. Most notable models, which focus on the less explored and challenging long-term time series forecasting (LTSF) problem, include LogTrans~\cite{li2019LogTrans} (NeurIPS 2019), Informer~\cite{informer} (AAAI 2021 Best paper), Autoformer~\cite{xu2021autoformer} (NeurIPS 2021), Pyraformer~\cite{liu2021pyraformer} (ICLR 2022 Oral), Triformer~\cite{cirstea2022triformer} (IJCAI 2022) and the recent FEDformer~\cite{zhou2022fedformer} (ICML 2022).

The main working power of Transformers is from its multi-head self-attention mechanism, which has a remarkable capability of extracting semantic correlations among elements in a long sequence (e.g., words in texts or 2D patches in images). However, self-attention is \emph{permutation-invariant} and ``anti-order'' to some extent. While using various types of positional encoding techniques can preserve some ordering information, it is still inevitable to have temporal information loss after applying self-attention on top of them. This is usually not a serious concern for semantic-rich applications such as NLP, e.g., the semantic meaning of a sentence is largely preserved even if we reorder some words in it. However, when analyzing time series data, there is usually a lack of semantics in the numerical data itself, and we are mainly interested in modeling the temporal changes among \emph{a continuous set of points}. That is, the order itself plays the most crucial role. Consequently, we pose the following intriguing question: \textbf{\emph{Are Transformers really effective for long-term time series forecasting?}}

Moreover, while existing Transformer-based LTSF solutions have demonstrated considerable prediction accuracy improvements over traditional methods, 
in their experiments, all the compared (non-Transformer) baselines perform autoregressive or iterated multi-step (IMS) forecasting~\cite{ariyo2014arima,DavidSalinas2017DeepARPF,DzmitryBahdanau2014NeuralMT,SeanJTaylor2017ForecastingAS}, which are known to suffer from significant error accumulation effects for the LTSF problem. Therefore, in this work, we challenge Transformer-based LTSF solutions with direct multi-step (DMS) forecasting strategies to validate their real performance.  

Not all time series are predictable, let alone long-term forecasting (e.g., for chaotic systems). We hypothesize that long-term forecasting is only feasible for those time series with a relatively clear trend and periodicity. As linear models can already extract such information, we introduce a set of embarrassingly simple models named \textbf{\modelname} as a new baseline for comparison. 
\modelname regresses historical time series with a one-layer linear model to forecast future time series directly. We conduct extensive experiments on nine widely-used benchmark datasets that cover various real-life applications: traffic, energy, economics, weather, and disease predictions.
Surprisingly, our results show that \modelname outperforms existing complex Transformer-based models \emph{in all cases, and often by a large margin (20\% $\sim$ 50\%)}. 
Moreover, we find that, in contrast to the claims in existing Transformers, most of them fail to extract temporal relations from long sequences, i.e., the forecasting errors are not reduced (sometimes even increased) with the increase of look-back window sizes. Finally, we conduct various ablation studies on existing Transformer-based TSF solutions to study the impact of various design elements in them. 

To sum up, the contributions of this work include:

\begin{itemize}

\item To the best of our knowledge, this is the first work to challenge the effectiveness of the booming Transformers for the long-term time series forecasting task. 

\item To validate our claims, we introduce a set of embarrassingly simple one-layer linear models, named \modelname, and compare them with existing Transformer-based LTSF solutions on nine benchmarks. \modelname can be a new baseline for the LTSF problem.

\item We conduct comprehensive empirical studies on various aspects of existing Transformer-based solutions, including the capability of modeling long inputs, the sensitivity to time series order, the impact of positional encoding and sub-series embedding, and efficiency comparisons. Our findings would benefit future research in this area. 

\end{itemize}

With the above, we conclude that \emph{the temporal modeling capabilities of Transformers for time series are exaggerated, at least for the existing LTSF benchmarks}. At the same time, while \modelname achieves a better prediction accuracy compared to existing works, it merely serves as a simple baseline for future research on the challenging long-term TSF problem. With our findings, we also advocate revisiting the validity of Transformer-based solutions for other time series analysis tasks in the future.

\section{Preliminaries: TSF Problem Formulation}
\label{sec:related}

\label{sec:pre_problem}
For time series containing $C$ variates, given historical data $\cX=\{X^t_{1}, ..., X^t_{C}\}{_{t=1}^{L}}$, wherein $L$ is the look-back window size and  $X^t_{i}$ is the value of the $i_{th}$ variate at the $t_{th}$ time step. The time series forecasting task is to predict the values $\hat{\cX}=\{\hat{X}^t_{1}, ..., \hat{X}^t_{C}\}{_{t=L+1}^{L+T}}$ at the $T$ future time steps. 
When $T>1$, iterated multi-step (IMS) forecasting~\cite{taieb2012recursive} learns a single-step forecaster and iteratively applies it to obtain multi-step predictions. Alternatively, direct multi-step (DMS) forecasting~\cite{chevillon2007direct} directly optimizes the multi-step forecasting objective at once.

Compared to DMS forecasting results, IMS predictions have smaller variance thanks to the autoregressive estimation procedure, but they inevitably suffer from error accumulation effects. Consequently, IMS forecasting is preferable when there is a highly-accurate single-step forecaster, and $T$ is relatively small. In contrast, DMS forecasting generates more accurate predictions when it is hard to obtain an unbiased single-step forecasting model, or $T$ is large.

\section{Transformer-Based LTSF Solutions}
\label{sec:trans}

\begin{figure*}[t]
\vspace{-0.2cm}
\begin{center}
\includegraphics[width=1\textwidth]{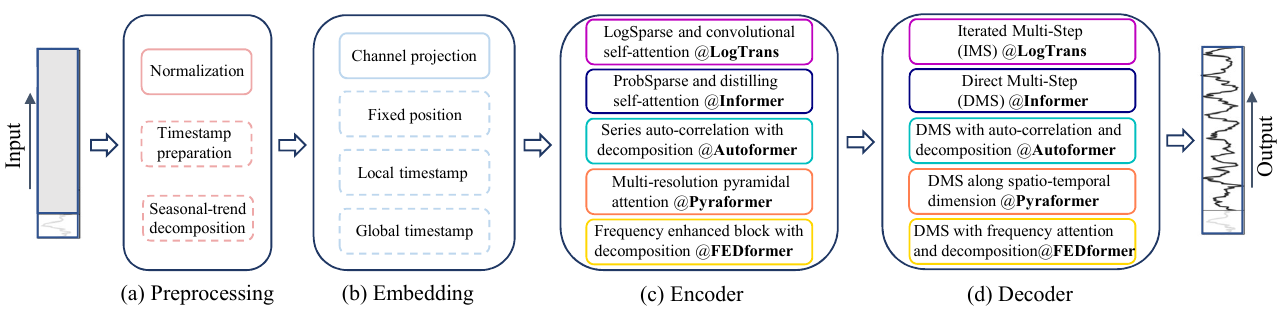}
\end{center}
\vspace{-0.4cm}
\caption{The pipeline of existing Transformer-based TSF solutions. In (a) and (b), the solid boxes are essential operations, and the dotted boxes are applied optionally. (c) and (d) are distinct for different methods~\cite{li2019LogTrans,informer,xu2021autoformer,liu2021pyraformer,zhou2022fedformer}.}
\vspace{-0.6cm}
\label{fig:pipeline}
\end{figure*}

Transformer-based models~\cite{vaswani2017attention} have achieved unparalleled performances in many long-standing AI tasks in natural language processing and computer vision fields, thanks to the effectiveness of the multi-head self-attention mechanism. This has also triggered lots of research interest in Transformer-based time series modeling techniques~\cite{wen2022transformers,minhao2021t}. In particular, a large amount of research works are dedicated to the LTSF task (e.g.,~\cite{li2019LogTrans,liu2021pyraformer,xu2021autoformer,informer,zhou2022fedformer}). Considering the ability to capture long-range dependencies with Transformer models, most of them focus on the less-explored long-term forecasting problem ($T\gg1$)\footnote{Due to page limit, we leave the discussion of non-Transformer forecasting solutions in the Appendix.}.

When applying the vanilla Transformer model to the LTSF problem, it has some limitations, including the quadratic time/memory complexity with the original self-attention scheme and error accumulation caused by the autoregressive decoder design. Informer~\cite{informer} addresses these issues and proposes a novel Transformer architecture with reduced complexity and a DMS forecasting strategy. Later, more Transformer variants 
introduce various time series features into their models for performance or efficiency improvements~\cite{liu2021pyraformer,xu2021autoformer,zhou2022fedformer}. We summarize the design elements of existing Transformer-based LTSF solutions as follows (see Figure~\ref{fig:pipeline}).

\noindent\textbf{Time series decomposition:}
For data preprocessing, normalization with zero-mean is common in TSF. Besides, Autoformer~\cite{xu2021autoformer} first applies seasonal-trend decomposition behind each neural block, which is a standard method in time series analysis to make raw data more predictable~\cite{RBCleveland1990STLA,hamilton2020time}. Specifically, they use a moving average kernel on the input sequence to extract the \emph{trend-cyclical} component of the time series. The difference between the original sequence and the trend component is regarded as the \emph{seasonal} component. 
On top of the decomposition scheme of Autoformer, FEDformer~\cite{zhou2022fedformer} further proposes the mixture of experts' strategies to mix the trend components extracted by moving average kernels with various kernel sizes.

\noindent\textbf{Input embedding strategies:}
The self-attention layer in the Transformer architecture cannot preserve the positional information of the time series. However, local positional information, i.e. the ordering of time series, is important. Besides, global temporal information, such as hierarchical timestamps (week, month, year) and agnostic timestamps (holidays and events), is also informative~\cite{informer}.
To enhance the temporal context of time-series inputs, a practical design in the SOTA Transformer-based methods is injecting several embeddings, like a fixed positional encoding, a channel projection embedding, and learnable temporal embeddings into the input sequence. Moreover, temporal embeddings with a temporal convolution layer~\cite{li2019LogTrans} or learnable timestamps~\cite{xu2021autoformer} are introduced. 

\noindent\textbf{Self-attention schemes:}
Transformers rely on the self-attention mechanism to extract the semantic dependencies between paired elements.
Motivated by reducing the $O\left(L^{2}\right)$ time and memory complexity of the vanilla Transformer, recent works propose two strategies for efficiency. On the one hand, LogTrans and Pyraformer explicitly introduce a sparsity bias into the self-attention scheme. Specifically, LogTrans uses a Logsparse mask to reduce the computational complexity to $O\left(LlogL\right)$ while Pyraformer adopts pyramidal attention that captures hierarchically multi-scale temporal dependencies with an $O\left(L\right)$ time and memory complexity. On the other hand, Informer and FEDformer use the low-rank property in the self-attention matrix. Informer proposes a ProbSparse self-attention mechanism and a self-attention distilling operation to decrease the complexity to $O\left(LlogL\right)$, and FEDformer designs a Fourier enhanced block and a wavelet enhanced block with random 
selection to obtain $O\left(L\right)$ complexity. Lastly, Autoformer designs a series-wise auto-correlation mechanism to replace the original self-attention layer.

\noindent\textbf{Decoders:}
The vanilla Transformer decoder outputs sequences in an autoregressive manner, resulting in a slow inference speed and error accumulation effects, especially for long-term predictions.
Informer designs a generative-style decoder for DMS forecasting. Other Transformer variants employ similar DMS strategies. 
For instance, Pyraformer uses a fully-connected layer concatenating Spatio-temporal axes as the decoder.
Autoformer sums up two refined decomposed features from trend-cyclical components and the stacked auto-correlation mechanism for seasonal components to get the final prediction. FEDformer also uses a decomposition scheme with the proposed frequency attention block to decode the final results.

The premise of Transformer models is the semantic correlations between paired elements, while the self-attention mechanism itself is permutation-invariant, and its capability of modeling temporal relations largely depends on positional encodings associated with input tokens. Considering the raw numerical data in time series (e.g., stock prices or electricity values), there are hardly any point-wise semantic correlations between them. In time series modeling, we are mainly interested in the temporal relations among a continuous set of points, and the order of these elements instead of the paired relationship plays the most crucial role. While employing positional encoding and using tokens to embed sub-series facilitate preserving some ordering information, the nature of the permutation-invariant self-attention mechanism inevitably results in temporal information loss. Due to the above observations, we are interested in revisiting the effectiveness of Transformer-based LTSF solutions.

\section{An Embarrassingly Simple Baseline}
\label{sec:method_simple}

In the experiments of existing Transformer-based LTSF solutions ($T\gg1$), all the compared (non-Transformer) baselines are IMS forecasting techniques, which are known to suffer from significant error accumulation effects. We hypothesize that the performance improvements in these works are largely due to the DMS strategy used in them. 

\begin{figure}[h]
\begin{center}
\includegraphics[width=0.32\textwidth]{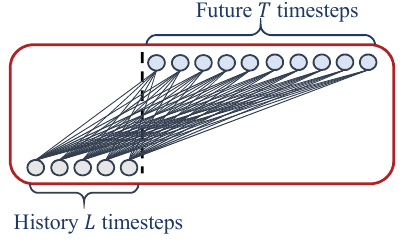}
\end{center}
\vspace{-0.3cm}
\caption{Illustration of the basic linear model.}
\vspace{-0.3cm}
\label{fig:Linear}
\end{figure}

To validate this hypothesis, we present the simplest DMS model via a temporal linear layer, named~\modelname, as a baseline for comparison. The basic formulation of \modelname directly regresses historical time series for future prediction via a weighted sum operation (as illustrated in Figure~\ref{fig:Linear}).
The mathematical expression is 
$\hat{X}_{i} = WX_i$, where $W\in \mathbb{R}^{T\times L}$ is a linear layer along the temporal axis. $\hat{X}_{i}$ and $X_i$ are the prediction and input for each $i_{th}$ variate. Note that \modelname shares weights across different variates and does not model any spatial correlations.

\modelname is a set of linear models. \textit{Vanilla Linear} is a one-layer linear model. To handle time series across different domains (e.g., finance, traffic, and energy domains), we further introduce two variants with two preprocessing methods, named \emph{DLinear} and \emph{NLinear}. 

\begin{itemize}

\item Specifically, \emph{DLinear} is a combination of a \emph{Decomposition} scheme used in Autoformer and FEDformer with linear layers. It first decomposes a raw data input into a trend component by a moving average kernel and a remainder (seasonal) component. Then, two one-layer linear layers are applied to each component, and we sum up the two features to get the final prediction.
By explicitly handling trend, \emph{DLinear} enhances the performance of a vanilla linear when there is a clear trend in the data. 
\item Meanwhile, to boost the performance of \modelname when there is a distribution shift in the dataset, \emph{NLinear} first subtracts the input by the last value of the sequence. Then, the input goes through a linear layer, and the subtracted part is added back before making the final prediction. The subtraction and addition in \emph{NLinear} are a simple normalization for the input sequence.
\end{itemize}

\section{Experiments}
\label{sec:exp}

\begin{table*}[h]
\vspace{-0.4cm}
\begin{center}
\scalebox{0.8}
{
\begin{tabular}{c|ccccccc} \hline
Datasets       & ETTh1$\&$ETTh2       & ETTm1 $\&$ETTm2        & Traffic     & Electricity  & Exchange-Rate & Weather      & ILI      \\ \hline
Variates       & 7          & 7            & 862         & 321          & 8             & 21          & 7      \\
Timesteps      & 17,420     & 69,680       & 17,544      & 26,304       & 7,588         & 52,696       & 966     \\
Granularity    & 1hour      & 5min         & 1hour       & 1hour        & 1day          & 10min        & 1week    \\ \hline
\end{tabular}}
\end{center}
\vspace{-0.5cm}
\caption{The statistics of the nine popular datasets for the LTSF problem.}
\label{tab:datasets}
\vspace{-0.2cm}
\end{table*}

\begin{table*}[t!]
\centering
\scalebox{0.7}{
\begin{tabular}{c|c|c|cccccc||cccccccccc|cc}
\hline
\multicolumn{2}{c|}{Methods}&{IMP.}&\multicolumn{2}{c|}{Linear*}&\multicolumn{2}{c|}{NLinear*}&\multicolumn{2}{c||}{DLinear*} &\multicolumn{2}{c|}{FEDformer}&\multicolumn{2}{c|}{Autoformer}&\multicolumn{2}{c|}{Informer}&\multicolumn{2}{c|}{Pyraformer*}&\multicolumn{2}{c|}{LogTrans}&\multicolumn{2}{c}{\emph{Repeat}*}\\
\hline
\multicolumn{2}{c|}{Metric} & MSE & MSE  & MAE& MSE  & MAE& MSE  & MAE& MSE  & MAE& MSE  & MAE& MSE  & MAE& MSE & MAE& MSE & MAE& MSE & MAE\\
\hline
\multirow{4}{*}{\rotatebox{90}{Electricity}}
&96&27.40\%&\textbf{0.140} & \textbf{0.237} & 0.141 & \textbf{0.237} & \textbf{0.140} & \textbf{0.237} & {\underline{0.193}} & \underline{0.308} & 0.201 & 0.317 & 0.274 & 0.368 & 0.386 & 0.449 & 0.258 & 0.357 & 1.588 & 0.946 \\
&192&23.88\%&\textbf{0.153} & 0.250 & 0.154 & \textbf{0.248} & \textbf{0.153} & {0.249} & \underline{0.201} & \underline{0.315} & 0.222 & 0.334 & 0.296 & 0.386 & 0.386 & 0.443 & 0.266 & 0.368 & 1.595 & 0.950 \\
&336&21.02\%&\textbf{0.169} & 0.268 & 0.171 & \textbf{0.265} & \textbf{0.169} & {0.267} & \underline{0.214} & \underline{0.329} & 0.231 & 0.338 & 0.300 & 0.394 & 0.378 & 0.443 & 0.280 & 0.380 & 1.617 & 0.961 \\
&720&17.47\%&\textbf{0.203} & {0.301} & {0.210} & \textbf{0.297} & \textbf{0.203} & {0.301} & \underline{0.246} & \underline{0.355} & 0.254 & 0.361 & 0.373 & 0.439 & 0.376 & 0.445 & 0.283 & 0.376 & 1.647 & 0.975 \\
\hline
\multirow{4}{*}{\rotatebox{90}{Exchange}}
&96&45.27\%&0.082 & 0.207 & 0.089 & 0.208 & \textbf{0.081} & {0.203} & \underline{0.148} & \underline{0.278} & 0.197 & 0.323 & 0.847 & 0.752 & 0.376 & 1.105 & 0.968 & 0.812 & {\textbf{0.081}} & {\textbf{0.196}} \\
&192&42.06\%&{0.167} & 0.304 & 0.180 & 0.300 & \textbf{0.157} & {0.293} & \underline{0.271} & \underline{0.380} & 0.300 & 0.369 & 1.204 & 0.895 & 1.748 & 1.151 & 1.040 & 0.851 & {0.167} & {\textbf{0.289}} \\
&336&33.69\%&0.328 & 0.432 & 0.331 & 0.415 & \textbf{0.305} & {0.414} & \underline{0.460} & \underline{0.500} & 0.509 & 0.524 & 1.672 & 1.036 & 1.874 & 1.172 & 1.659 & 1.081 & \textbf{0.305} & \textbf{0.396} \\
&720&46.19\%&0.964 & 0.750 & 1.033 & 0.780 & \textbf{0.643} & \textbf{0.601} & \underline{1.195} & \underline{0.841} & 1.447 & 0.941 & 2.478 & 1.310 & 1.943 & 1.206 & 1.941 & 1.127 & {0.823} & {0.681} \\
\hline
\multirow{4}{*}{\rotatebox{90}{Traffic}}
&96&30.15\%&\textbf{0.410} & {0.282} & \textbf{0.410} & \textbf{0.279} & \textbf{0.410} & {0.282} & {\underline{0.587}} & {\underline{0.366}} & {{0.613}} & 0.388 & 0.719 & 0.391 & 2.085 & 0.468 & 0.684 & {{0.384}} & 2.723 & 1.079 \\
&192&29.96\%&\textbf{0.423} & {0.287} & \textbf{0.423} & \textbf{0.284} & \textbf{0.423} & {0.287} & \underline{0.604} & {\underline{0.373}} & 0.616 & 0.382 & 0.696 & 0.379 & 0.867 & 0.467 & 0.685 & 0.390 & 2.756 & 1.087 \\
&336&29.95\%&{0.436} & {0.295} & \textbf{0.435} & \textbf{0.290} & 0.436 & 0.296 & \underline{0.621} & {0.383} & {0.622} & {\underline{0.337}} & 0.777 & 0.420 & 0.869 & 0.469 & 0.734 & 0.408 & 2.791 & 1.095 \\
&720&25.87\%&{0.466} & {0.315} & \textbf{0.464} & \textbf{0.307} & {0.466} & {0.315} & {\underline{0.626}} & {\underline{0.382}} & 0.660 & 0.408 & 0.864 & 0.472 & 0.881 & 0.473 & 0.717 & 0.396 & 2.811 & 1.097 \\
\hline
\multirow{4}{*}{\rotatebox{90}{Weather}}
&96&18.89\%&\textbf{0.176} & {0.236} & 0.182 & \textbf{0.232} & \textbf{0.176} & 0.237 & \underline{0.217} & \underline{0.296} & 0.266 & 0.336 & 0.300 & 0.384 & 0.896 & 0.556 & 0.458 & 0.490 & 0.259 & {{0.254}} \\
&192&21.01\%&\textbf{0.218} & {0.276} & 0.225 & \textbf{0.269} & 0.220 & 0.282 &\underline{0.276} & \underline{0.336} & 0.307 & 0.367 & 0.598 & 0.544 & 0.622 & 0.624 & 0.658 & 0.589 & 0.309 & {{0.292}} \\
&336&22.71\%&\textbf{0.262} & {0.312} & 0.271 & \textbf{0.301} & 0.265 & 0.319 & \underline{0.339} & \underline{0.380} & 0.359 & 0.395 & 0.578 & 0.523 & 0.739 & 0.753 & 0.797 & 0.652 & 0.377 & 0.338 \\
&720&19.85\%&0.326 & 0.365 & 0.338 & \textbf{0.348} & \textbf{0.323} & {0.362} & \underline{0.403} & \underline{0.428} & 0.419 & 0.428 & 1.059 & 0.741 & 1.004 & 0.934 & 0.869 & 0.675 & 0.465 & 0.394 \\
\hline
\multirow{4}{*}{\rotatebox{90}{ILI}}
&24&47.86\%&{1.947} & {0.985} & \textbf{1.683} & \textbf{0.858} & 2.215 & 1.081 & \underline{3.228} & \underline{1.260} & 3.483 & 1.287 & 5.764 & 1.677 & 1.420 & 2.012 & 4.480 & 1.444 & 6.587 & 1.701 \\
&36&36.43\%&2.182 & 1.036 & \textbf{1.703} & \textbf{0.859} & {1.963} & {0.963} & {\underline{2.679}} & {\underline{1.080}} & 3.103 & 1.148 & 4.755 & 1.467 & 7.394 & 2.031 & 4.799 & 1.467 & 7.130 & 1.884 \\
&48&34.43\%&2.256 & 1.060 & \textbf{1.719} & \textbf{0.884} & {2.130} & {1.024} & \underline{2.622} & {\underline{1.078}} & 2.669 & 1.085 & 4.763 & 1.469 & 7.551 & 2.057 & 4.800 & 1.468 & 6.575 & 1.798 \\
&60&34.33\%&2.390 & 1.104 & \textbf{1.819} & \textbf{0.917} & {2.368} & {1.096} & {2.857} & {1.157} & {\underline{2.770}} & {\underline{1.125}} & 5.264 & 1.564 & 7.662 & 2.100 & 5.278 & 1.560 & 5.893 & 1.677 \\\hline
\multirow{4}{*}{\rotatebox{90}{ETTh1}}
 & 96 &0.80\% & {0.375} & { 0.397} & \textbf{0.374} & \textbf{0.394} & {  {0.375}} & { {0.399}}  & { \underline{0.376}} &{ \underline{0.419}}        & 0.449 & 0.459 & 0.865 & 0.713 & 0.664 & 0.612 & 0.878 & 0.740  &1.295	&0.713\\
 & 192  &3.57\% & 0.418 & 0.429 & {0.408} & \textbf{0.415} & { \textbf{0.405}} & {  {0.416}}  & \underline{  0.420} &{ \underline{0.448}}       & 0.500 & {0.482} & {1.008}  & 0.792 & 0.790 & 0.681 & 1.037 & 0.824  &1.325	&0.733\\
 & 336 &6.54\% & 0.479 & 0.476 & \textbf{0.429} & \textbf{0.427} & {  {0.439}} & {  {0.443}}   & { \underline{0.459}} &{ \underline{0.465}}       & 0.521 & 0.496 & 1.107 & 0.809 & 0.891 & 0.738 & 1.238 & 0.932  &1.323	&0.744\\
 & 720 &13.04\%& 0.624 & 0.592 & \textbf{0.440} & \textbf{0.453} & {  {0.472}} & {  {0.490}}        & { \underline{0.506}}    & { \underline{0.507}} &{ {0.514}} &{ {0.512}} & 1.181 & 0.865 & 0.963 & 0.782 & 1.135 & 0.852  &1.339	&0.756\\
\hline
\multirow{4}{*}{\rotatebox{90}{ETTh2}}
 & 96  & 19.94\%&  {0.288} &  {0.352}  & \textbf{0.277} & \textbf{0.338} & { {0.289}} & { {0.353}}          &{\underline{0.346}}    &{ \underline{0.388}}              & 0.358       & 0.397       & 3.755 & 1.525 & 0.645 & 0.597 & 2.116 & 1.197 &0.432	&0.422 \\
 & 192 & 19.81\%&  {0.377} &  {0.413}  &\textbf{0.344} & \textbf{0.381} & { {0.383}} & { {0.418}}                 & { \underline{0.429}} & { \underline{0.439}}    & 0.456       &{ {0.452}}       & 5.602 & 1.931 & 0.788 & 0.683 & 4.315 & 1.635 &0.534	&0.473 \\
 & 336 & 25.93\% & 0.452 & 0.461  & \textbf{0.357} & \textbf{0.400} & {  {0.448}} & {  {0.465}}                &{ \underline{0.496}}          &{ \underline{0.487}} & { {0.482}} & { {0.486}} & 4.721 & 1.835 & 0.907 & 0.747 & 1.124 & 1.604 &0.591	&0.508\\
 & 720 & 14.25\%& 0.698 & 0.595  & \textbf{0.394} & \textbf{0.436} & 0.605 & 0.551            & {  \underline{0.463}} & {  \underline{0.474} }&{ {0.515}}       &{ {0.511}}       & 3.647 & 1.625 & 0.963 & 0.783 & 3.188 & 1.540  &0.588	&0.517\\
\hline
\multirow{4}{*}{\rotatebox{90}{ETTm1}}
 & 96  & 21.10\% & 0.308 & 0.352  &  {0.306} &  {0.348} & { \textbf{0.299}} & { \textbf{0.343}} &{ \underline{0.379}}          &{ \underline{0.419} }        & 0.505       & 0.475       & 0.672 & 0.571 & 0.543 & 0.510 & 0.600 & 0.546  &1.214	&0.665\\
 & 192 & 21.36\% & {0.340} & {0.369}  & 0.349 & 0.375 & { \textbf{0.335}} & { \textbf{0.365}} &{ \underline{0.426}}          &{ \underline{0.441}}       & 0.553       & 0.496       & 0.795 & 0.669 & 0.557 & 0.537 & 0.837 & 0.700  &1.261	&0.690\\
 & 336 & 17.07\% & 0.376 & 0.393  & {0.375} & {0.388} & {\textbf {0.369}} & { \textbf{0.386}} &{ \underline{0.445}}          &{ \underline{0.459}}     & 0.621       & 0.537       & 1.212 & 0.871 & 0.754 & 0.655 & 1.124 & 0.832  &1.283	&0.707\\
 & 720 &21.73\% & 0.440 & 0.435  & {0.433} & {0.422} & { \textbf{0.425}} & { \textbf{0.421}} &{ \underline{0.543}}          &{ \underline{0.490} }     & 0.671       & 0.561       & 1.166 & 0.823 & 0.908 & 0.724 & 1.153 & 0.820  &1.319	&0.729\\
\hline
\multirow{4}{*}{\rotatebox{90}{ETTm2}} 
 & 96  & 17.73\% & 0.168 & 0.262  &\textbf{ 0.167} & \textbf{0.255} & { \textbf{0.167}} & { {0.260}} &{ \underline{0.203}}    &{\underline{0.287}}    & 0.255       & 0.339       & 0.365 & 0.453 & 0.435 & 0.507 & 0.768 & 0.642 & 0.266	& 0.328\\
 & 192 & 17.84\% & 0.232 & 0.308  & \textbf{0.221} & \textbf{0.293} & { {0.224}} & { {0.303}}      &{ \underline{0.269}}    &{\underline{0.328}}    & 0.281       & 0.340       & 0.533 & 0.563 & 0.730 & 0.673 & 0.989 & 0.757  & 0.340	& 0.371\\
 & 336 & 15.69\% & 0.320 & 0.373  & \textbf{0.274} & \textbf{0.327} & { {0.281}}    & { {0.342}}       & { \underline{0.325}} & { \underline{0.366}}  & 0.339       & 0.372       & 1.363 & 0.887 & 1.201 & 0.845 & 1.334 & 0.872  & 0.412	& 0.410\\
 & 720 & 12.58\% & 0.413 & 0.435 & \textbf{0.368} & \textbf{0.384} & { {0.397}}    & 0.421        & { \underline{0.421}} & { \underline{0.415}} & 0.433         & 0.432          & 3.379 & 1.338 & 3.625 & 1.451 & 3.048 & 1.328 & 0.521 &0.465\\
\hline
\end{tabular}
}
\begin{tablenotes} 
    \tiny
    {
    \item - Methods* are implemented by us; Other results are from FEDformer~\cite{zhou2022fedformer}.
    }
\end{tablenotes} 
\vspace{-0.2cm}
\caption{Multivariate long-term forecasting errors in terms of MSE and MAE, the lower the better. Among them, ILI dataset is with forecasting horizon $T \in \{24,36,48,60\}$. For the others, $T \in \{96,192,336,720\}$. \emph{Repeat} repeats the last value in the look-back window. The \textbf{best results} are highlighted in {\textbf{bold}} and the \underline{best results of Transformers} are highlighted with a { \underline{underline}}. Accordingly, IMP. is the best result of linear models compared to the results of Transformer-based solutions. }
\vspace{-0.5cm}
\label{tab:multi-benchmarks-ett}
\end{table*}

\begin{figure*}[h]	
	\subfigure[Electricity] 
	{
		\begin{minipage}{5cm}
			\centering          
			\includegraphics[scale=0.23]{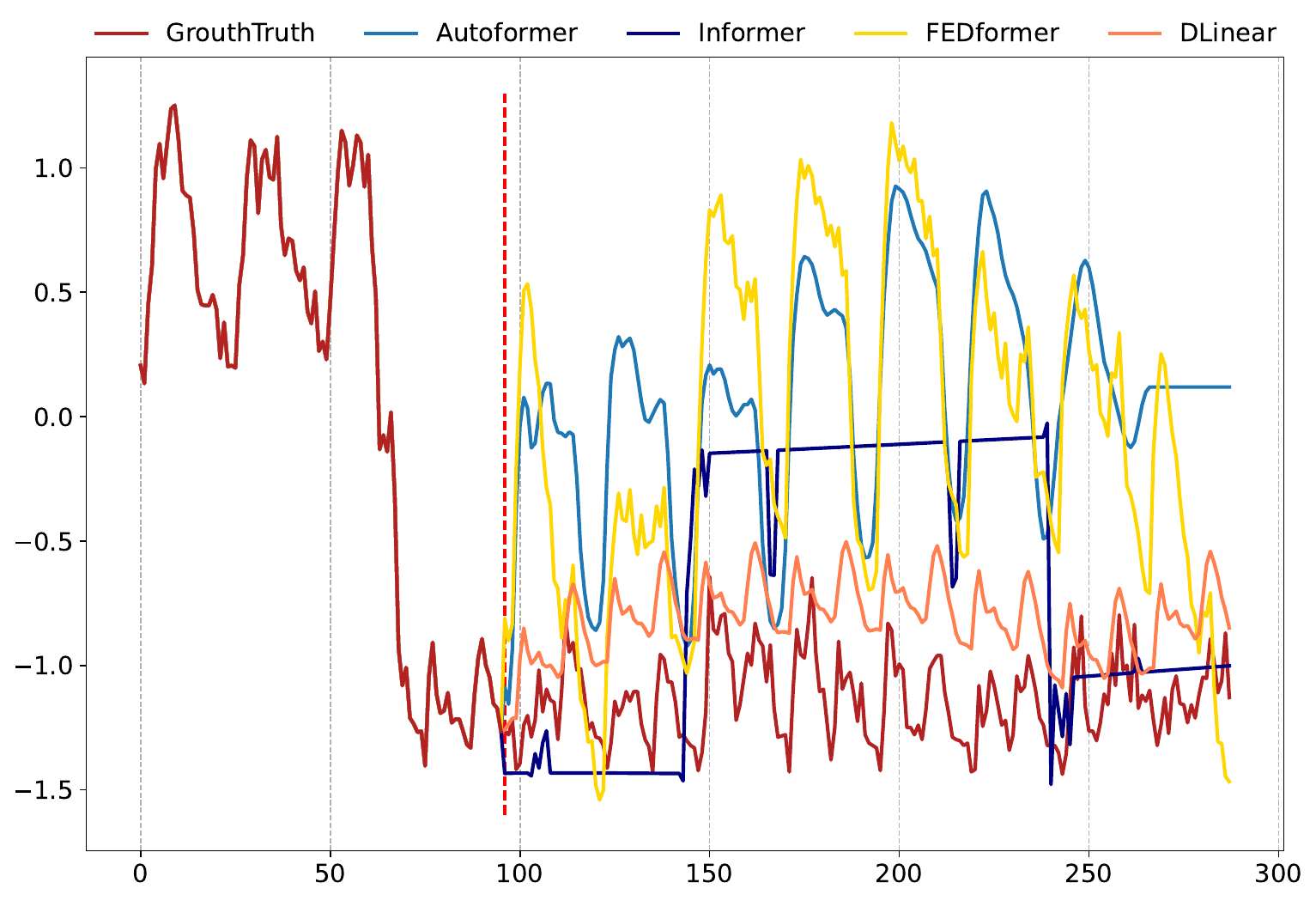}   
		\end{minipage}
	}
	\hspace{0.55cm}
	\subfigure[Exchange-Rate] 
	{
		\begin{minipage}{5cm}
			\centering          
			\includegraphics[scale=0.23]{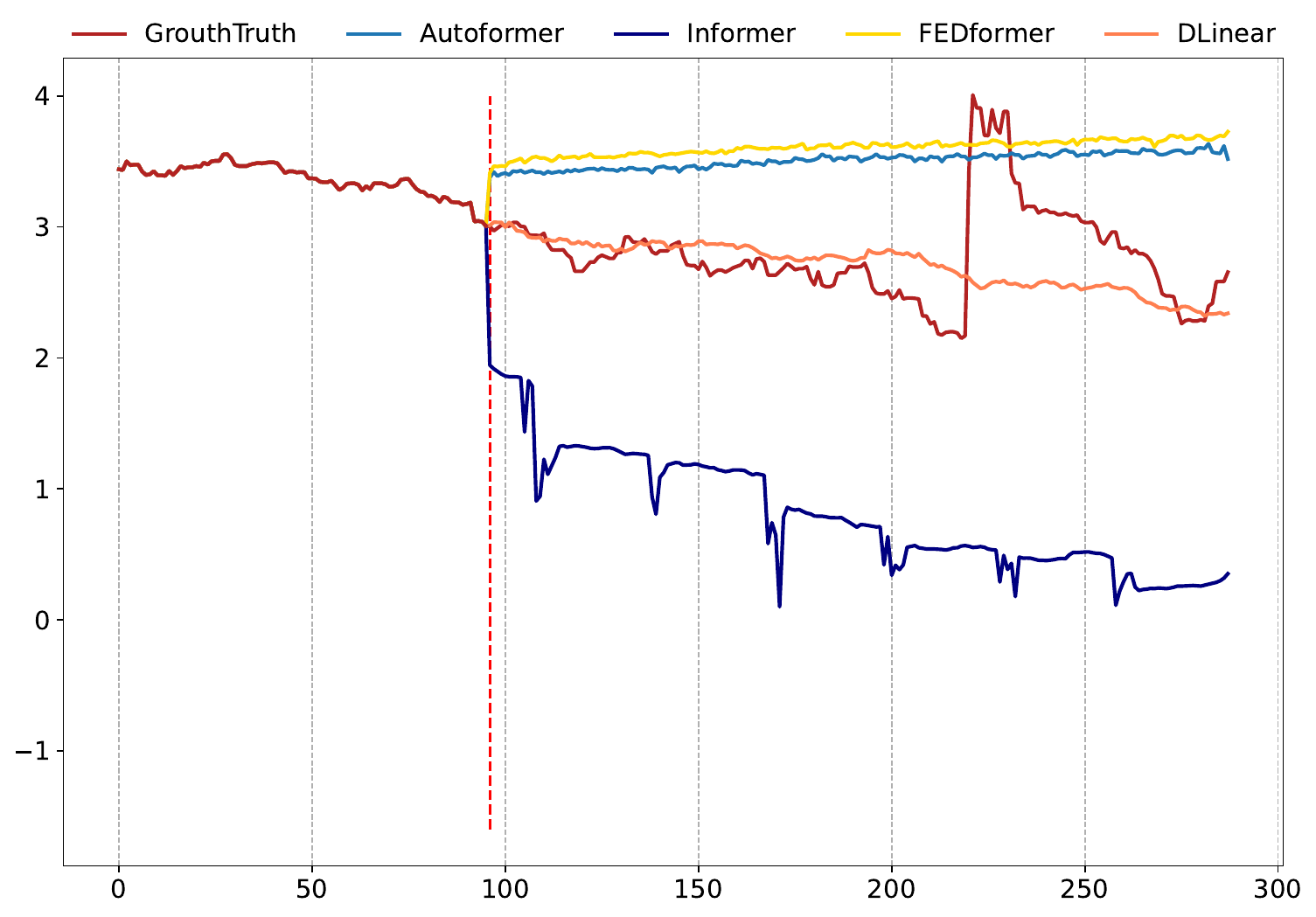}   
		\end{minipage}
	}
	\hspace{0.45cm}
	\subfigure[ETTh2] 
	{
		\begin{minipage}{5cm}
			\centering      
			\includegraphics[scale=0.23]{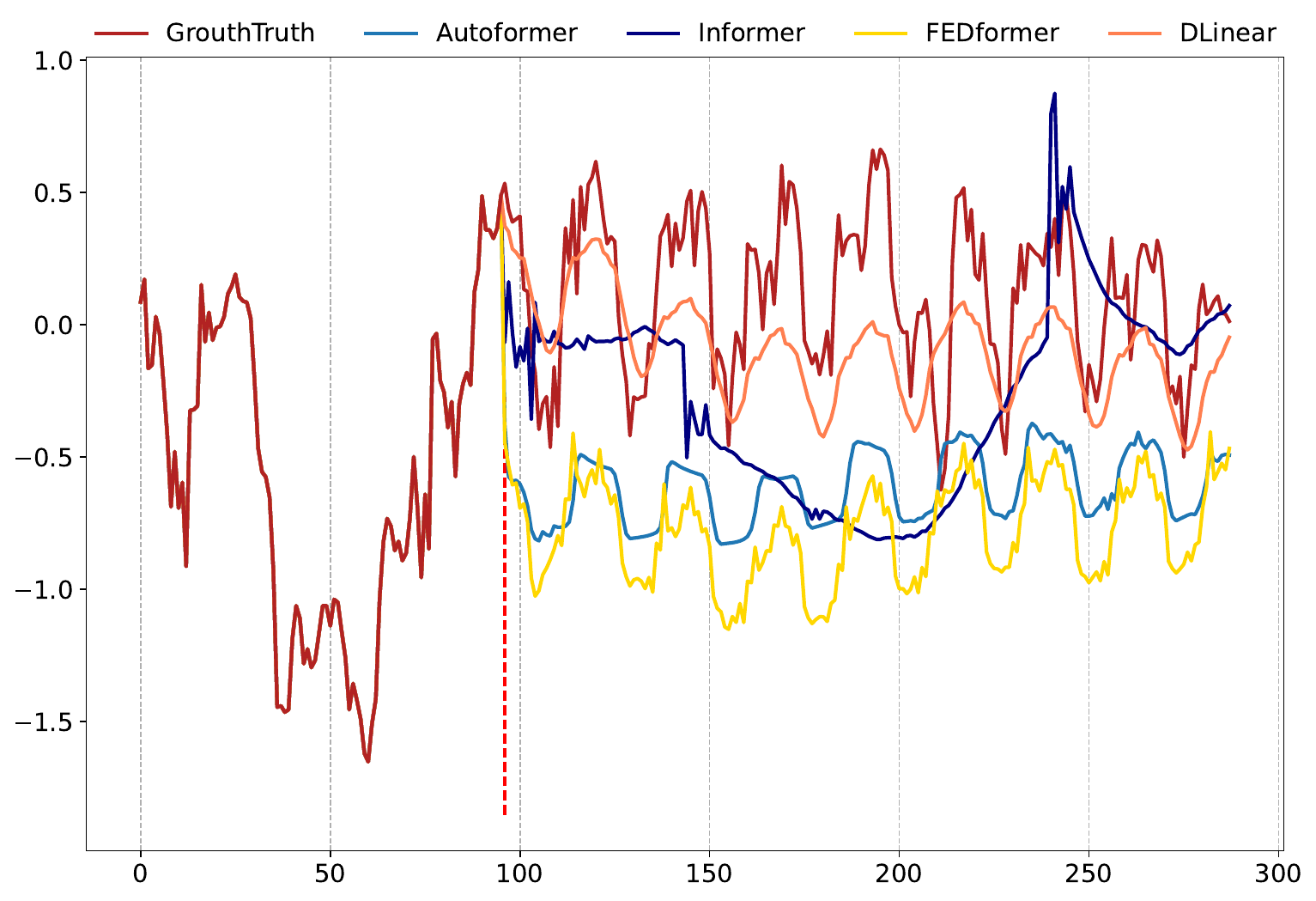}   
		\end{minipage}
	}
    \vspace{-0.3cm}
	\caption{Illustration of the long-term forecasting output (Y-axis) of five models with an input length $L$=96 and output length $T$=192 (X-axis) on Electricity, Exchange-Rate, and ETTh2, respectively.} %
	\label{fig:visualize_prediction}  
\vspace{-0.2cm}
\end{figure*}

\subsection{Experimental Settings}

\textbf{Dataset.} We conduct extensive experiments on nine widely-used real-world datasets, including ETT (Electricity Transformer Temperature)~\cite{informer} (ETTh1, ETTh2, ETTm1, ETTm2), Traffic, Electricity, Weather, ILI, Exchange-Rate~\cite{GuokunLai2017lstm}. All of them are multivariate time series. We leave \emph{data descriptions} in the Appendix.

\textbf{Evaluation metric.}
Following previous works~\cite{informer,xu2021autoformer,zhou2022fedformer}, we use Mean Squared Error (MSE) and Mean Absolute Error (MAE) as the core metrics to compare performance.

\textbf{Compared methods.} We include five recent Transformer-based methods: FEDformer~\cite{zhou2022fedformer}, Autoformer~\cite{xu2021autoformer}, Informer~\cite{informer}, Pyraformer~\cite{liu2021pyraformer}, and LogTrans~\cite{li2019LogTrans}. 
Besides, we include a naive DMS method: Closest Repeat (\emph{Repeat}), which repeats the last value in the look-back window, as another simple baseline. Since there are two variants of FEDformer, we compare the one with better accuracy (FEDformer-f via Fourier transform). 

\subsection{Comparison with Transformers}

\textbf{Quantitative results.} In Table~\ref{tab:multi-benchmarks-ett}, we extensively evaluate all mentioned Transformers on nine benchmarks, following the experimental setting of previous work~\cite{xu2021autoformer,zhou2022fedformer,informer}.
Surprisingly, the performance of \modelname surpasses the SOTA FEDformer in most cases by 20\% $\sim$ 50\% improvements on the \emph{multivariate forecasting}, where \modelname even does not model correlations among variates. For different time series benchmarks, \emph{NLinear} and \emph{DLinear} show the superiority to handle the distribution shift and trend-seasonality features.
We also provide results for \emph{univariate forecasting} of ETT datasets in the Appendix, where \modelname still consistently outperforms Transformer-based LTSF solutions by a large margin.

FEDformer achieves competitive forecasting accuracy on ETTh1. This because FEDformer employs classical time series analysis techniques such as frequency processing, which brings in time series inductive bias and benefits the ability of temporal feature extraction. 
In summary, these results reveal that existing complex Transformer-based LTSF solutions are not seemingly effective on the existing nine benchmarks while \modelname can be a powerful baseline.

Another interesting observation is that even though the naive \emph{Repeat} method shows worse results when predicting long-term seasonal data (e.g., Electricity and Traffic), it surprisingly outperforms all Transformer-based methods on Exchange-Rate (around 45\%). 
This is mainly caused by the wrong prediction of trends in Transformer-based solutions, which may overfit toward sudden change noises in the training data, resulting in significant accuracy degradation (see Figure~\ref{fig:visualize_prediction}(b)). Instead, \emph{Repeat} does not have the bias.

\textbf{Qualitative results.}
As shown in Figure~\ref{fig:visualize_prediction}, we plot the prediction results on three selected time series datasets with Transformer-based solutions and \modelname: Electricity (Sequence 1951, Variate 36), Exchange-Rate (Sequence 676, Variate 3), and ETTh2 ( Sequence 1241, Variate 2), where these datasets have different temporal patterns. When the input length is 96 steps, and the output horizon is 336 steps, Transformers~\cite{informer,xu2021autoformer,zhou2022fedformer} fail to capture the scale and bias of the future data on Electricity and ETTh2. Moreover, they can hardly predict a proper trend on aperiodic data such as Exchange-Rate. These phenomena further indicate the inadequacy of existing Transformer-based solutions for the LTSF task.

\subsection{More Analyses on LTSF-Transformers}
\label{subsec:TransformerModelDesign}

\textbf{\emph{Can existing LTSF-Transformers extract temporal relations well from longer input sequences?}}
The size of the look-back window greatly impacts forecasting accuracy as it determines how much we can learn from historical data. Generally speaking, a powerful TSF model with a strong temporal relation extraction capability should be able to achieve better results with larger look-back window sizes.

\begin{figure}[h]	
    \hspace{-0.6cm}
	\subfigure[\textbf{720} steps-\textbf{Traffic}]
	{
		\begin{minipage}{3.5cm}
			\centering     
			\includegraphics[scale=0.3]{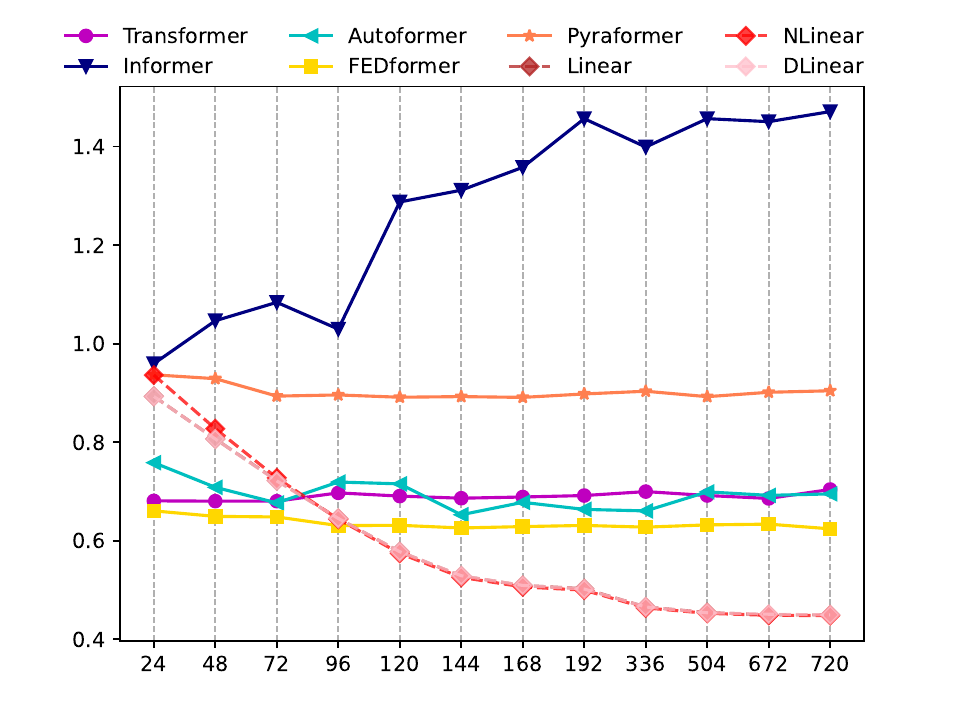}
		\end{minipage}
	}
    \hspace{0.6cm}
	\subfigure[\textbf{720} steps-\textbf{Electricity}]
	{
		\begin{minipage}{3.5cm}
			\centering     
			\includegraphics[scale=0.3]{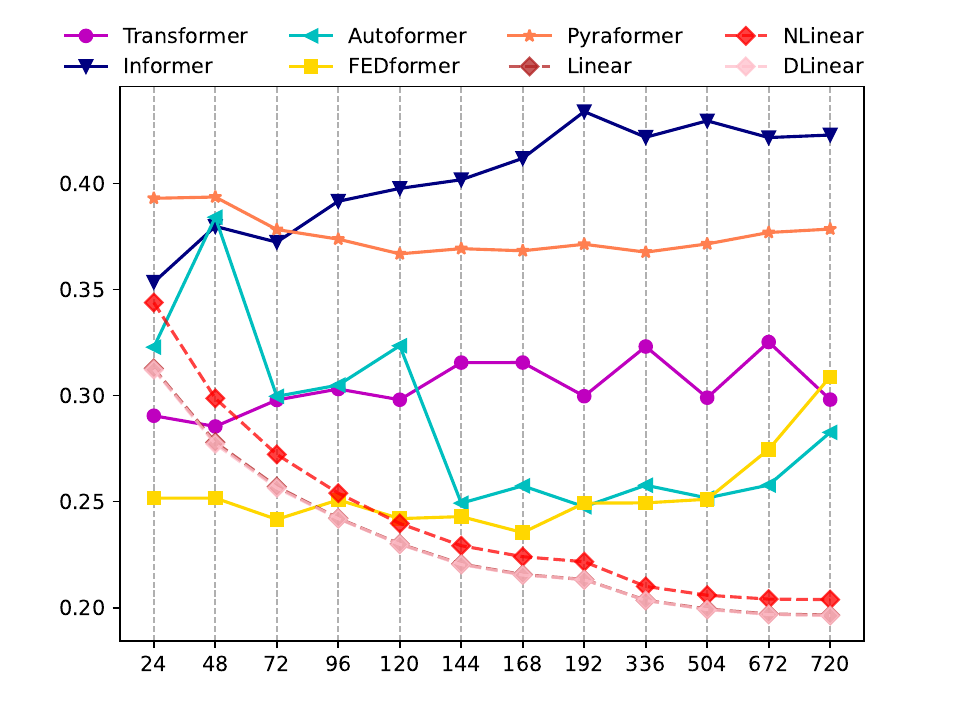} 
		\end{minipage}
	}
    \vspace{-0.2cm}
	\caption{The MSE results (Y-axis) of models with different look-back window sizes (X-axis) of long-term forecasting (T=\textbf{720}) on the Traffic and Electricity datasets. } 
	\label{fig:lookBackWindows}  
\vspace{-0.2cm}
\end{figure}

To study the impact of input look-back window sizes, we conduct experiments with $L \in \{24, 48, 72, 96, 120, 144, 168, 192, 336, 504, 672, 720\}$ for long-term forecasting (T=720).
Figure~\ref{fig:lookBackWindows} demonstrates the MSE results on two datasets. 
Similar to the observations from previous studies~\cite{informer,wen2022transformers}, existing Transformer-based models' performance deteriorates or stays stable when the look-back window size increases. In contrast, the performances of all \modelname are significantly boosted with the increase of look-back window size. 
Thus, existing solutions tend to overfit temporal noises instead of extracting temporal information if given a longer sequence, and the input size 96 is exactly suitable for most Transformers.
Additionally, we provide more quantitative results in the Appendix, and our conclusion holds in almost all cases. 

\textbf{\emph{What can be learned for long-term forecasting?}}
While the temporal dynamics in the look-back window significantly impact the forecasting accuracy of short-term time series forecasting, we hypothesize that long-term forecasting depends on whether \emph{models can capture the trend and periodicity well only.} That is, the farther the forecasting horizon, the less impact the look-back window itself has.  
\begin{table}[h]
\centering
\scalebox{0.9}{
\begin{tabular}{c|cc|cc}
\hline
      \multicolumn{1}{c|}{Methods} &\multicolumn{2}{c|}{FEDformer} & \multicolumn{2}{c}{Autoformer}   \\ \hline
        Input & \emph{Close} & \emph{Far} & \emph{Close} & \emph{Far}  \\ \hline
        Electricity  
        & 0.251 & 0.265 & 0.255 & 0.287  \\
        Traffic 
       & 0.631 & 0.645 & 0.677 & 0.675  \\\hline
        
\end{tabular}
}
\vspace{-0.2cm}
\caption{Comparison of different input sequences under the MSE metric to explore what LTSF-Transformers depend on. If the input is \emph{Close}, we use the $96_{th},..., 191_{th}$ time steps as the input sequence. If the input is \emph{Far}, we use the $0_{th},..., 95_{th}$ time steps. Both of them forecast the $192_{th},...,(192+720)_{th}$ time steps. }
\label{tab:Prior}
\vspace{-0.4cm}
\end{table}

To validate the above hypothesis, in Table~\ref{tab:Prior}, we compare the forecasting accuracy for the same future 720 time steps with data from two different look-back windows: (i). the original input L=96 setting (called \emph{Close}) and (ii). the far input L=96 setting (called \emph{Far}) that is before the original 96 time steps. From the experimental results, the performance of the SOTA Transformers drops slightly, indicating these models only capture similar temporal information from the adjacent time series sequence. Since capturing the intrinsic characteristics of the dataset generally does not require a large number of parameters, i,e. one parameter can represent the periodicity. Using too many parameters will even cause overfitting, which partially explains why \modelname performs better than Transformer-based methods.

\textbf{\emph{Are the self-attention scheme effective for LTSF?}}
We verify whether these complex designs in the existing Transformer (e.g., Informer) are essential. In Table \ref{tab:Attention}, we gradually transform Informer to Linear. First, we replace each self-attention layer by a linear layer, called \emph{Att.-Linear}, since a self-attention layer can be regarded as a fully-connected layer where weights are dynamically changed.
Furthermore, we discard other auxiliary designs (e.g., FFN) in Informer to leave embedding layers and linear layers, named \emph{Embed + Linear}. Finally, we simplify the model to one linear layer. Surprisingly, the performance of Informer grows with the gradual simplification, indicating the unnecessary of the self-attention scheme and other complex modules at least for existing LTSF benchmarks.

\begin{table}[h]
\centering
\scalebox{0.85}{
\begin{tabular}{c|c|cccc}
\hline
        \multicolumn{2}{c|}{Methods} & Informer & \emph{Att.-Linear} & \emph{Embed + Linear} & Linear  \\ \hline
        \multirow{4}{*}{\rotatebox{90}{Exchange}} 
        & 96 & 0.847 & 1.003 & 0.173 & 0.084  \\ 
        & 192 & 1.204 & 0.979 & 0.443 & 0.155  \\ 
        & 336 & 1.672 & 1.498 & 1.288 & 0.301  \\ 
        & 720 & 2.478 & 2.102 & 2.026 & 0.763  \\ \hline
        \multirow{4}{*}{\rotatebox{90}{ETTh1}} 
        & 96 & 0.865 & 0.613 & 0.454 & 0.400  \\ 
        & 192 & 1.008 & 0.759 & 0.686 & 0.438  \\ 
        & 336 & 1.107 & 0.921 & 0.821 & 0.479  \\ 
        & 720 & 1.181 & 0.902 & 1.051 & 0.515 \\ \hline
    \end{tabular}
}
\vspace{-0.2cm}
\caption{The MSE comparisons of gradually transforming Informer to a Linear from the left to right columns. \emph{Att.-Linear} is a structure that replaces each attention layer with a linear layer. \emph{Embed + Linear} is to drop other designs and only keeps embedding layers and a linear layer. The look-back window size is 96.}
\label{tab:Attention}
\vspace{-0.3cm}
\end{table}

\begin{table*}[h]
\centering
\scalebox{0.85}{
\begin{tabular}{c|c|ccc|ccc|ccc|ccc}
\hline
\multicolumn{2}{c|}{Methods} & \multicolumn{3}{c|}{Linear} &\multicolumn{3}{c|}{FEDformer} & \multicolumn{3}{c|}{Autoformer} & \multicolumn{3}{c}{Informer} \\ \hline
\multicolumn{2}{c|}{Predict Length} & \emph{Ori.} & \emph{Shuf.} & \emph{Half-Ex.} & \emph{Ori.} & \emph{Shuf.} & \emph{Half-Ex.} & \emph{Ori.} & \emph{Shuf.} & \emph{Half-Ex.} & \emph{Ori.} & \emph{Shuf.} & \emph{Half-Ex.} \\ \hline
        \multirow{4}{*}{\rotatebox{90}{Exchange}}
        &96 & 0.080 & 0.133 & 0.169 & 0.161 & 0.160 & 0.162 & 0.152 & 0.158 & 0.160 & 0.952 & 1.004 & 0.959  \\ 
        &192 & 0.162 & 0.208 & 0.243 & 0.274 & 0.275 & 0.275 & 0.278 & 0.271 & 0.277 & 1.012 & 1.023 & 1.014 \\ 
        &336 & 0.286 & 0.320 & 0.345 & 0.439 & 0.439 & 0.439 & 0.435 & 0.430 & 0.435 & 1.177 & 1.181 & 1.177  \\ 
        &720 & 0.806 & 0.819 & 0.836 & 1.122 & 1.122 & 1.122 & 1.113 & 1.113 & 1.113 & 1.198 & 1.210 & 1.196  \\ \hline
        &Average Drop & N/A & 27.26\% & 46.81\% & N/A & -0.09\% & 0.20\% & N/A & 0.09\% & 1.12\% & N/A & -0.12\% & -0.18\%\\ \hline
        \multirow{4}{*}{\rotatebox{90}{ETTh1}}
        &96 & 0.395 & 0.824 & 0.431 & 0.376 & 0.753 & 0.405 & 0.455 & 0.838 & 0.458 & 0.974 & 0.971 & 0.971   \\ 
        &192 & 0.447 & 0.824 & 0.471 & 0.419 & 0.730 & 0.436 & 0.486 & 0.774 & 0.491 & 1.233 & 1.232 & 1.231   \\ 
        &336 & 0.490 & 0.825 & 0.505 & 0.447 & 0.736 & 0.453 & 0.496 & 0.752 & 0.497 & 1.693 & 1.693 & 1.691 \\ 
        &720 & 0.520 & 0.846 & 0.528 & 0.468 & 0.720 & 0.470 & 0.525 & 0.696 & 0.524 & 2.720 & 2.716 & 2.715  \\ \hline
        &Average Drop & N/A & 81.06\% & 4.78\% & N/A & 73.28\% & 3.44\% & N/A & 56.91\% & 0.46\% & N/A & 1.98\% & 0.18\%  \\ \hline
\end{tabular}
}
\vspace{-0.2cm}
\caption{The MSE comparisons of models when shuffling the raw input sequence. \emph{Shuf.} randomly shuffles the input sequence. \emph{Half-EX.} randomly exchanges the first half of the input sequences with the second half. Average Drop is the average performance drop under all forecasting lengths after shuffling. All results are the average test MSE of five runs.}
\label{tab:Shuffle}
\vspace{-0.4cm}
\end{table*}

\textbf{\emph{Can existing LTSF-Transformers preserve temporal order well?}}
Self-attention is inherently permutation-invariant, i.e., regardless of the order. However, in time-series forecasting, the sequence order often plays a crucial role. We argue that even with positional and temporal embeddings, existing Transformer-based methods still suffer from temporal information loss. In Table ~\ref{tab:Shuffle}, we shuffle the raw input before the embedding strategies. Two shuffling strategies are presented: \emph{Shuf.} randomly shuffles the whole input sequences and \emph{Half-Ex.} exchanges the first half of the input sequence with the second half.
Interestingly, compared with the original setting (\emph{Ori.}) on the Exchange Rate, the performance of all Transformer-based methods does not fluctuate even when the input sequence is randomly shuffled. By contrary, the performance of \modelname is damaged significantly. These indicate that LTSF-Transformers with different positional and temporal embeddings preserve quite limited temporal relations and are prone to overfit on noisy financial data, while the \modelname can model the order naturally and avoid overfitting with fewer parameters. 

For the ETTh1 dataset, FEDformer and Autoformer introduce time series inductive bias into their models, making them can extract certain temporal information when the dataset has more clear temporal patterns (e.g., periodicity) than the Exchange Rate. Therefore, the average drops of the two Transformers are 73.28\% and 56.91\% under the \emph{Shuf.} setting, where it loses the whole order information. Moreover, Informer still suffers less from both \emph{Shuf.} and \emph{Half-Ex.} settings due to its no such temporal inductive bias. Overall, the average drops of \modelname are larger than Transformer-based methods for all cases, indicating the existing Transformers do not preserve temporal order well.

%

\begin{table}[h]
    \centering
    \scalebox{0.85}
    {
            \begin{tabular}{c|c|cccc}
            \hline
            \multirow{2}{*}{Methods}    & \multirow{2}{*}{Embedding}                & \multicolumn{4}{c}{Traffic}                                   \\
                                         &                            & 96             & 192            & 336            & 720            \\\hline \hline
            \multirow{4}{*}{FEDformer}   & All                       & 0.597          & 0.606          & 0.627          & 0.649          \\
                                         & wo/Pos.                   & \textbf{0.587} & \textbf{0.604} & \textbf{0.621} & \textbf{0.626} \\
                                         & wo/Temp.                  & 0.613          & 0.623          & 0.650          & 0.677          \\
                                         & wo/Pos.-Temp.             & 0.613          & 0.622          & 0.648          & 0.663          \\\hline\hline
            \multirow{4}{*}{Autoformer}  & All              & 0.629          & 0.647          & 0.676          & \textbf{0.638} \\
                                         & wo/Pos.          & \textbf{0.613} & \textbf{0.616} & \textbf{0.622} & 0.660          \\
                                         & wo/Temp.         & 0.681          & 0.665          & 0.908          & 0.769          \\
                                         & wo/Pos.-Temp.    & 0.672          & 0.811          & 1.133          & 1.300          \\\hline\hline
            \multirow{4}{*}{Informer}    & All              & \textbf{0.719} & \textbf{0.696} & \textbf{0.777} & \textbf{0.864}          \\
                                         & wo/Pos.          & 1.035          & 1.186          & 1.307          & 1.472          \\
                                         & wo/Temp.         & 0.754          & 0.780          & 0.903          & 1.259 \\
                                         & wo/Pos.-Temp.    & 1.038          & 1.351          & 1.491          & 1.512         
             \\\hline
            \end{tabular}
            }
            \vspace{-0.3cm}
            \caption{The MSE comparisons of different embedding strategies on Transformer-based methods with look-back window size $96$ and forecasting lengths $\{96,192,336,720\}$. }
\label{tab:embedding}
\vspace{-0.5cm}
\end{table}

\textbf{\emph{How effective are different embedding strategies?}}
We study the benefits of position and timestamp embeddings used in Transformer-based methods. 
%
In Table~\ref{tab:embedding}, the forecasting errors of Informer largely increase without positional embeddings (wo/Pos.). Without timestamp embeddings (wo/Temp.) will gradually damage the performance of Informer as the forecasting lengths increase. 
Since Informer uses a single time step for each token, it is necessary to introduce temporal information in tokens.

Rather than using a single time step in each token, FEDformer and Autoformer input a sequence of timestamps to embed the temporal information. 
Hence, they can achieve comparable or even better performance without fixed positional embeddings. However, without timestamp embeddings, the performance of Autoformer declines rapidly because of the loss of global temporal information. Instead, thanks to the frequency-enhanced module proposed in FEDformer to introduce temporal inductive bias, it suffers less from removing any position/timestamp embeddings. 

\textbf{\emph{Is training data size a limiting factor for existing LTSF-Transformers?}}
Some may argue that the poor performance of Transformer-based solutions is due to the small sizes of the benchmark datasets. 
Unlike computer vision or natural language processing tasks, TSF is performed on collected time series, and it is difficult to scale up the training data size. In fact, the size of the training data would indeed have a significant impact on the model performance. Accordingly, we conduct experiments on Traffic, comparing the performance of the model trained on a full dataset (17,544*0.7 hours), named \emph{Ori.}, with that trained on a shortened dataset (8,760 hours, i.e., 1 year), called \emph{Short}. Unexpectedly, Table~\ref{tab:shortenTraffic} presents that the prediction errors with reduced training data are lower in most cases. This might because the whole-year data maintains more clear temporal features than a longer but incomplete data size. While we cannot conclude that we should use less data for training, it demonstrates that the training data scale is not the limiting reason for the performances of Autoformer and FEDformer. 

\begin{table}[h]
\centering
\scalebox{0.9}{
\begin{tabular}{c|cc|cc}
\hline
Methods        & \multicolumn{2}{c|}{FEDformer}     & \multicolumn{2}{c}{Autoformer}    \\ \hline
Dataset& \emph{Ori.} & \emph{Short} & \emph{Ori.} & \emph{Short}  \\\hline
96                  &0.587 & \textbf{0.568} &0.613 & \textbf{0.594} \\
192                  &0.604 & \textbf{0.584} &\textbf{0.616} & 0.621 \\
336                  &0.621 & \textbf{0.601} &0.622 & \textbf{0.621} \\
720                   &0.626 & \textbf{0.608} &0.660 & \textbf{0.650} \\ \hline
\end{tabular}
}
\vspace{-0.2cm}
\caption{The MSE comparison of two training data sizes.}
\vspace{-0.3cm}
\label{tab:shortenTraffic}
\end{table}

\textbf{\emph{Is efficiency really a top-level priority?}}
Existing LTSF-Transformers claim that the $O\left(L^{2}\right)$ complexity of the vanilla Transformer is unaffordable for the LTSF problem. Although they prove to be able to improve the theoretical time and memory complexity from $O\left(L^{2}\right)$ to $O\left(L\right)$, it is unclear whether \emph{1) the actual inference time and memory cost on devices are improved, and 2) the memory issue is unacceptable and urgent for today's GPU (e.g., an NVIDIA Titan XP here).}
In Table ~\ref{tab:efficiency}, we compare the average practical efficiencies with 5 runs. Interestingly, compared with the vanilla Transformer (with the same DMS decoder), most Transformer variants incur similar or even worse inference time and parameters in practice. These follow-ups introduce more additional design elements to make practical costs high. Moreover, the memory cost of the vanilla Transformer is practically acceptable, even for output length $L=720$, which weakens the importance of developing a memory-efficient Transformers, at least for existing benchmarks.

\begin{table}[h]
\vspace{-0.1cm}
        \centering
        \scalebox{0.9}
        {
        	{%
                \begin{tabular}{c|cccc} \hline
                Method      & MACs & Parameter & Time & Memory\\\hline
                DLinear  & \textbf{0.04G}   & \textbf{139.7K}    & \textbf{0.4ms}     &\textbf{687MiB}        \\\hline
                Transformer$\times$ & 4.03G   & 13.61M     & 26.8ms     &6091MiB \\
                Informer    & 3.93G   & 14.39M     & 49.3ms     &3869MiB        \\
                Autoformer  & 4.41G   & 14.91M     & 164.1ms     &7607MiB        \\
                Pyraformer & 0.80G    &  241.4M$^*$   & 3.4ms     &7017MiB          \\
                FEDformer   & 4.41G   & 20.68M     & 40.5ms     &4143MiB \\\hline       
                \end{tabular}%
        	}
            }
            \begin{tablenotes}
            \tiny
            {
            \item - $\times$ is modified into the same one-step decoder, which is implemented in the source code from Autoformer.
            \item - $*$ 236.7M parameters of Pyraformer come from its linear decoder. 
            }
            \end{tablenotes} 
            \vspace{-0.2cm}
        \caption{Comparison of practical efficiency of LTSF-Transformers under L=96 and T=720 on the Electricity. MACs are the number of multiply-accumulate operations. We use Dlinear for comparison since it has the double cost in \modelname. The inference time averages 5 runs.}
\label{tab:efficiency}
\vspace{-0.3cm}
\end{table}

\vspace{-0.4cm}
\section{Conclusion and Future Work}
\label{sec:conclusion}
\textbf{Conclusion.}
This work questions the effectiveness of emerging favored Transformer-based solutions for the long-term time series forecasting problem. We use an embarrassingly simple linear model~\modelname as a DMS forecasting baseline to verify our claims. Note that our contributions do not come from proposing a linear model but rather from throwing out an important question, showing surprising comparisons, and demonstrating why LTSF-Transformers are not as effective as claimed in these works through various perspectives.
We sincerely hope our comprehensive studies can benefit future work in this area.

\textbf{Future work.} \modelname has a limited model capacity, and it merely serves a simple yet competitive baseline with strong interpretability for future research. For example, the one-layer linear network is hard to capture the temporal dynamics caused by change points~\cite{van2020evaluation}. Consequently, we believe there is a great potential for new model designs, data processing, and benchmarks to tackle the challenging LTSF problem. 

\pagebreak

\twocolumn[{
	\renewcommand\twocolumn[1][]{#1}
	\begin{center}
		\textbf{\Large Appendix:\\Are Transformers Effective for Time Series Forecasting?}\end{center}
}]

In this Appendix, we provide descriptions of non-Transformer-based TSF solutions, detailed experimental settings, more comparisons under different look-back window sizes, and the visualization of \modelname on all datasets. We also append our code to reproduce the results shown in the paper.

\appendix

\section{Related Work: Non-Transformer-Based TSF Solutions}
\label{sec:related_non}
As a long-standing problem with a wide range of applications, statistical approaches (e.g., autoregressive integrated moving average (ARIMA)~\cite{ariyo2014arima}, exponential smoothing~\cite{gardner1985exponential}, and structural models~\cite{harvey1990forecasting}) for time series forecasting have been used from the 1970s onward.
Generally speaking, the parametric models used in statistical methods require significant domain expertise to build. 

To relieve this burden, many machine learning techniques such as gradient boosting regression tree (GBRT)~\cite{gbrt,elsayed2021we} 
gain popularity, which learns the temporal dynamics of time series in a data-driven manner. 
However, these methods still require manual feature engineering and model designs.
With the powerful representation learning capability of deep neural networks (DNNs) from abundant data, various deep learning-based TSF solutions are proposed in the literature, achieving better forecasting accuracy than traditional techniques in many cases. 

Besides Transformers, the other two popular DNN architectures are also applied for time series forecasting: 

\begin{itemize}
    \item Recurrent neural networks (RNNs) based methods (e.g.,~\cite{petnehazi2019recurrent}) summarize the past information compactly in internal memory states and recursively update themselves for forecasting. 
    \item Convolutional neural networks (CNNs) based methods (e.g.,~\cite{bai2018empirical}), wherein convolutional filters are used to capture local temporal features.
\end{itemize}

RNN-based TSF methods belong to IMS forecasting techniques. Depending on whether the decoder is implemented in an autoregressive manner, there are either IMS or DMS forecasting techniques for CNN-based TSF methods~\cite{bai2018empirical,liu2021time}. 

\section{Experimental Details}
\label{sec:supp_exp}
\subsection{Data Descriptions}
We use nine wildly-used datasets in the main paper. The details are listed in the following.
\begin{itemize}
\item ETT (Electricity Transformer Temperature)~\cite{informer}\footnote{\url{https://github.com/zhouhaoyi/ETDataset}} consists of two hourly-level datasets (ETTh) and two 15-minute-level datasets (ETTm). Each of them contains seven oil and load features of electricity transformers from July $2016$ to July $2018$.
\item Traffic\footnote{\url{http://pems.dot.ca.gov}} describes the road occupancy rates. It contains the hourly data recorded by the sensors of San Francisco freeways from $2015$ to $2016$. 
\item Electricity\footnote{\url{https://archive.ics.uci.edu/ml/datasets/ElectricityLoadDiagrams20112014}} collects the hourly electricity consumption of $321$ clients from $2012$ to $2014$.
\item Exchange-Rate~\cite{GuokunLai2017lstm}\footnote{\url{https://github.com/laiguokun/multivariate-time-series-data}} collects the daily exchange rates of $8$ countries from $1990$ to $2016$.
\item Weather\footnote{\url{https://www.bgc-jena.mpg.de/wetter/}} includes $21$ indicators of weather, such as air temperature, and humidity. Its data is recorded every 10 min for $2020$ in Germany.
\item ILI\footnote{\url{https://gis.cdc.gov/grasp/fluview/fluportaldashboard.html}} describes the ratio of patients seen with influenza-like illness and the number of patients. It includes weekly data from the Centers for Disease Control and Prevention of the United States from $2002$ to $2021$.

\end{itemize}

\subsection{Implementation Details}
For existing Transformer-based TSF solutions: the implementation of Autoformer~\cite{xu2021autoformer}, Informer~\cite{informer}, and the vanilla Transformer~\cite{vaswani2017attention} are all taken from the Autoformer work~\cite{xu2021autoformer}; the implementation of FEDformer~\cite{zhou2022fedformer} and Pyraformer~\cite{liu2021pyraformer} are from their respective code repository. We also adopt their default hyper-parameters to train the models. For \emph{DLinear}, the moving average kernel size for decomposition is 25, which is the same as Autoformer. The total parameters of a vanilla linear model and a \emph{NLinear} are T\*L. The total parameters of the \emph{DLinear} are 2\*T\*L. Since \modelname will be underfitting when the input length is short, and LTSF-Transformers tend to overfit on a long lookback window size. To compare the best performance of existing LTSF-Transformers with \modelname, we report L=336 for \modelname and L=96 for Transformers by default. For more hyper-parameters of \modelname, please refer to our code.

\section{Additional Comparison with Transformers}
\label{sec:supp_add}
We further compare \modelname with LTSF-Transformer for  Univariate Forecasting on four ETT datasets. Moreover, in Figure $4$ of the main paper, we demonstrate that existing Transformers fail to exploit large look-back window sizes with two examples. Here, we give comprehensive comparisons between \modelname and Transformer-based TSF solutions under various look-back window sizes \emph{on all benchmarks}.

\subsection{Comparison of Univariate Forecasting}

\begin{figure*}[h]	
	\subfigure[ETTh1 channel6]
	{
		\begin{minipage}{4.1cm}
			\centering     
			\includegraphics[scale=0.29]{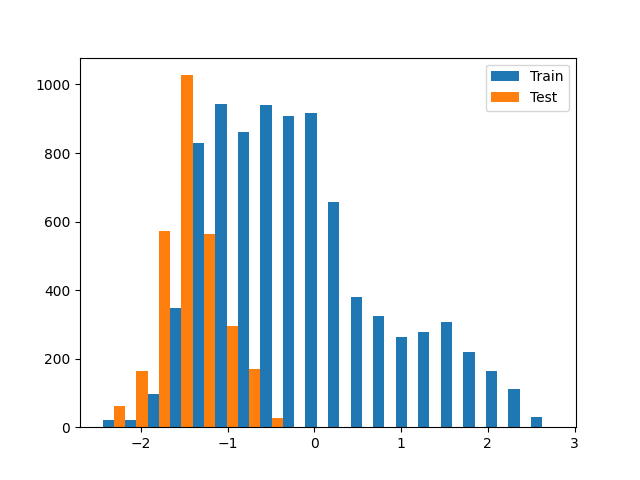}
		\end{minipage}
	}
	\subfigure[ETTh2 channel3]
	{
		\begin{minipage}{4.1cm}
			\centering     
			\includegraphics[scale=0.29]{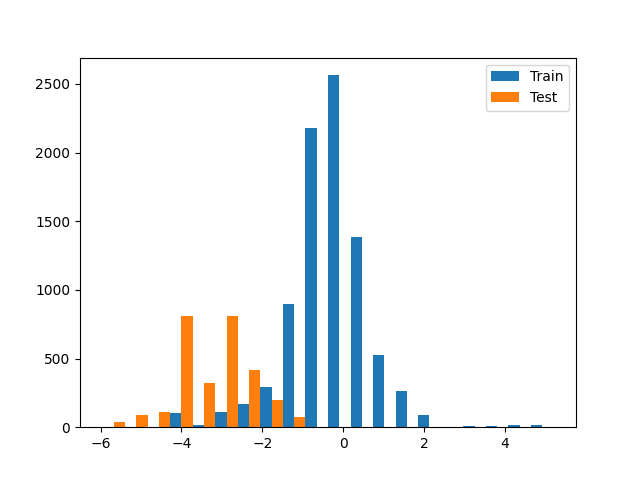} 
		\end{minipage}
	}
	\subfigure[Electricity channel3]
	{
		\begin{minipage}{4.1cm}
			\centering     
			\includegraphics[scale=0.29]{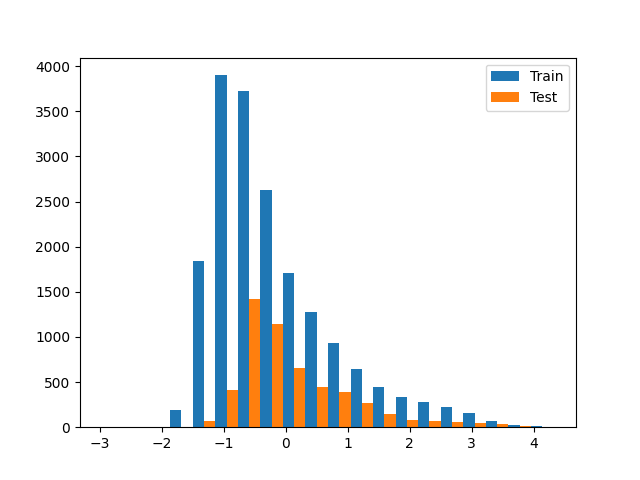} 
		\end{minipage}
	}
	\subfigure[ILI channel6] 
	{
		\begin{minipage}{4.1cm}
			\centering         
			\includegraphics[scale=0.29]{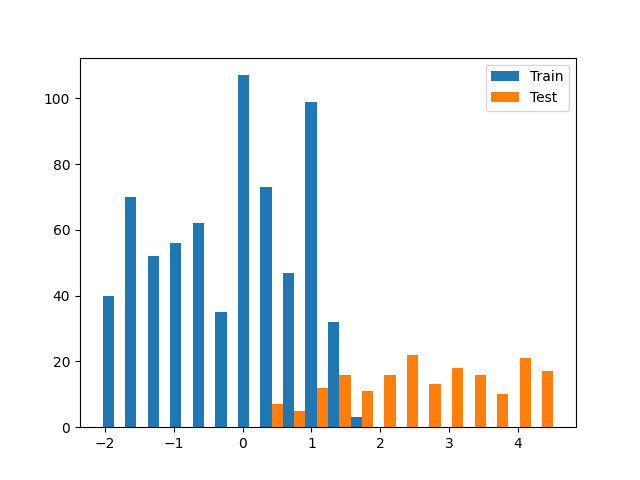} 
		\end{minipage}
	}
\caption{Distribution of ETTh1, ETTh2, Electricity, and ILI dataset. A clear distribution shift between training and testing data can be observed in ETTh1, ETTh2, and ILI. } 
	\label{fig:distribution_shift} 
\end{figure*}

We present the univariate forecasting results on the four ETT datasets in table~\ref{tab:uni-benchmarks-ett}. Similarly, \modelname, especially for \emph{NLinear} can consistently outperform all transformer-based methods by a large margin in most time. We find that there are serious distribution shifts between training and test sets (as shown in Fig.~\ref{fig:distribution_shift} (a), (b)) on ETTh1 and ETTh2 datasets. Simply normalization via the last value from the lookback window can greatly relieve the distribution shift problem.

\begin{table*}[h]
\centering
\scalebox{0.85}{
\begin{tabular}{c|c|cccccc|ccccccccccc}
\hline
\multicolumn{2}{c|}{Methods}&\multicolumn{2}{c|}{Linear}&\multicolumn{2}{c|}{NLinear}&\multicolumn{2}{c|}{DLinear}&\multicolumn{2}{c|}{FEDformer-f}&\multicolumn{2}{c|}{FEDformer-w}&\multicolumn{2}{c|}{Autoformer}&\multicolumn{2}{c|}{Informer}&\multicolumn{2}{c}{LogTrans}\\
\hline
\multicolumn{2}{c|}{Metric} & MSE  & MAE & MSE  & MSE  & MAE & MAE & MSE & MAE& MSE  & MAE& MSE  & MAE& MSE  & MAE& MSE  & MAE\\
\hline

\multirow{4}{*}{\rotatebox{90}{$ETTh1$}}
&96 &0.189	&0.359	&\textbf{0.053}	&\textbf{0.177}	& {0.056} &{0.180} &0.079 &0.215 &0.080 &0.214 &{\underline{0.071}} &{\underline{0.206}}&	0.193&	0.377&	0.283&	0.468\\
&192 &0.078	&0.212	&\textbf{0.069}	&\textbf{0.204}	& {0.071}	&{0.204} &{\underline{0.104}} &{\underline{0.245}} &0.105 &0.256 & 0.114 &	0.262&	0.217&	0.395&	0.234&	0.409\\
&336 &0.091	&0.237	&\textbf{0.081}	&\textbf{0.226}	& {0.098}	&{0.244} &0.119 &0.270 &0.120 &0.269 &{\underline{0.107}}&{\underline{0.258}} &0.202&	0.381&	0.386&	0.546\\
&720 &0.172	&0.340	&\textbf{0.080}	&\textbf{0.226}	&0.189	&0.359 &0.142 &0.299 & {0.127} &{0.280} &{\underline{0.126}}&{\underline{0.283}} &0.183&	0.355&	0.475&	0.629\\
\hline
\multirow{4}{*}{\rotatebox{90}{$ETTh2$}}
&96 &0.133	&0.283	&\textbf{0.129}	&\textbf{0.278}	&0.131	&0.279 &{\underline{0.128}} &{\underline{0.271}} &0.156 &0.306 & 0.153&	0.306&	0.213&	0.373&	0.217&	0.379\\
&192 &0.176	&0.330	&\textbf{0.169}	&\textbf{0.324}	&{0.176}	&{0.329} &{\underline{0.185}} &{\underline{0.330}} &0.238 &0.380 & 0.204&	0.351&	0.227&	0.387&	0.281&	0.429\\
&336 &0.213	&0.371	&\textbf{0.194}	&\textbf{0.355}	&{0.209}	&{0.367} &{\underline{0.231}} &{\underline{0.378}} &0.271 &0.412 & 0.246&	0.389&	0.242&	0.401&	0.293&	0.437\\
&720 &0.292	&0.440	&\textbf{0.225}	&\textbf{0.381}	&0.276	&0.426 &0.278 &{{0.420}} &0.288 &0.438 & {\underline{0.268}}&	{\underline{0.409}}&	0.291&	0.439&	0.218&	0.387\\
\hline
\multirow{4}{*}{\rotatebox{90}{$ETTm1$}}
&96 &0.028	&0.125	&\textbf{0.026}	&\textbf{0.122}	&{0.028}	&{0.123} &{\underline{0.033}} &{\underline{0.140}} &0.036 &0.149 & 0.056&	0.183&	0.109&	0.277&	0.049&	0.171\\
&192 &0.043	&0.154	&\textbf{0.039}	&\textbf{0.149}	&{0.045}	&{0.156} &{\underline{0.058}} &{\underline{0.186}} &0.069 &0.206 & 0.081&	0.216&	0.151&	0.310&	0.157&	0.317\\
&336 &0.059	&0.180	&\textbf{0.052}	&\textbf{0.172}	&{0.061}	&{0.182} &0.084 &0.231 &{\underline{0.071}} &{\underline{0.209}} & 0.076&	0.218&	0.427&	0.591&	0.289&	0.459\\
&720 &0.080	&0.211	&\textbf{0.073}	&\textbf{0.207}	&{0.080}	&{0.210} &{\underline{0.102}} &0.250 &0.105 &{\underline{0.248}} & 0.110&	0.267&	0.438&	0.586&	0.430&	0.579\\
\hline
\multirow{4}{*}{\rotatebox{90}{$ETTm2$}} 
& 96 &0.066	&0.189	&\textbf{0.063}	&\textbf{0.182}	&{0.063}	&{0.183} & 0.067 & 0.198 &{\underline{0.063}} &{\underline{0.189}} & 0.065 & 0.189 & 0.088 & 0.225 & 0.075 & 0.208   \\
& 192 &0.094	&0.230	&\textbf{0.090}	&\textbf{0.223}	&{0.092}	&{0.227} & {\underline{0.102}} & {\underline{0.245} }&0.110 &0.252 & 0.118 & 0.256 & 0.132 &0.283 & 0.129 &0.275     \\
& 336 &0.120	&0.263	&\textbf{0.117}	&\textbf{0.259}	&{0.119}	&{0.261} & {\underline{0.130}} & {\underline{0.279} }&0.147 &0.301 & 0.154 & 0.305  &0.180 &0.336 & 0.154 &0.302      \\
& 720 &0.175	&0.320	&\textbf{0.170}	&\textbf{0.318}	&{0.175}	&{0.320}& {\underline{0.178}} & {\underline{0.325}} &0.219 &0.368 & 0.182 &0.335  &0.300  &0.435 & 0.160 &0.321       \\
\hline
\end{tabular}
}
\caption{Univariate long sequence time-series forecasting results on ETT full benchmark. The \textbf{best results} are highlighted in { \textbf{bold}} and the \underline{best results of Transformers} are highlighted with a { \underline{underline}}.}
\label{tab:uni-benchmarks-ett}
\end{table*}

\begin{figure*}[h]	
	\subfigure[\textbf{24} steps-\textbf{ETTh1}]
	{
		\begin{minipage}{4.1cm}
			\centering     
			\includegraphics[scale=0.3]{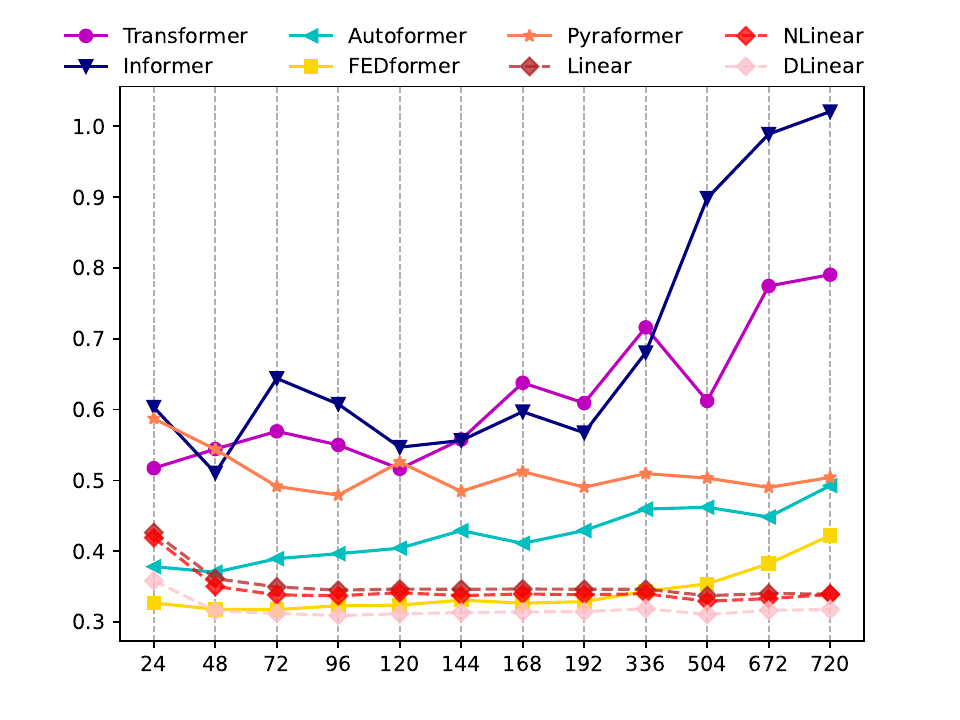}
		\end{minipage}
	}
	\subfigure[\textbf{720} steps-\textbf{ETTh1}]
	{
		\begin{minipage}{4.1cm}
			\centering     
			\includegraphics[scale=0.3]{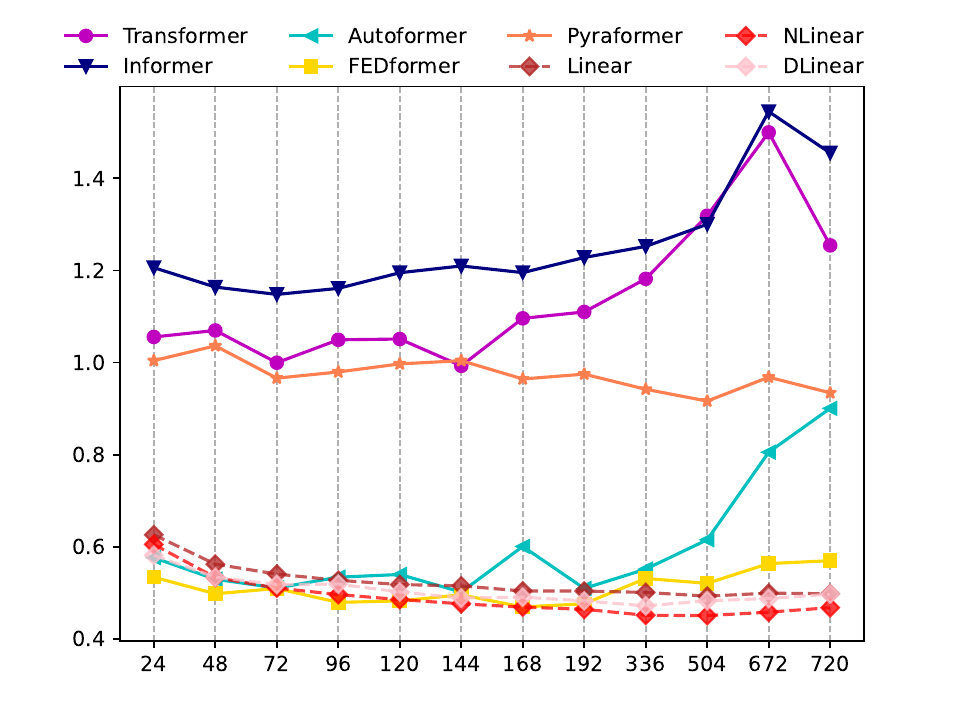} 
		\end{minipage}
	}
	\hspace{-2pt}
	\subfigure[\textbf{24} steps-\textbf{ETTh2}] 
	{
		\begin{minipage}{4.1cm}
			\centering         
			\includegraphics[scale=0.3]{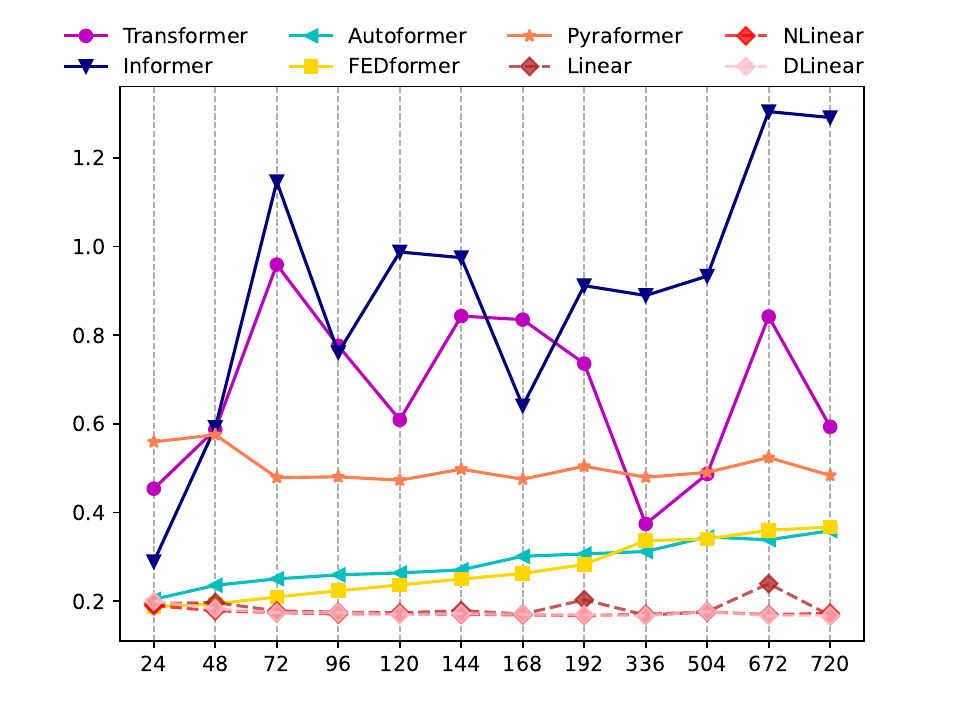} 
		\end{minipage}
	}
	\subfigure[\textbf{720} steps-\textbf{ETTh2}] 
	{
		\begin{minipage}{4.1cm}
			\centering      
			\includegraphics[scale=0.3]{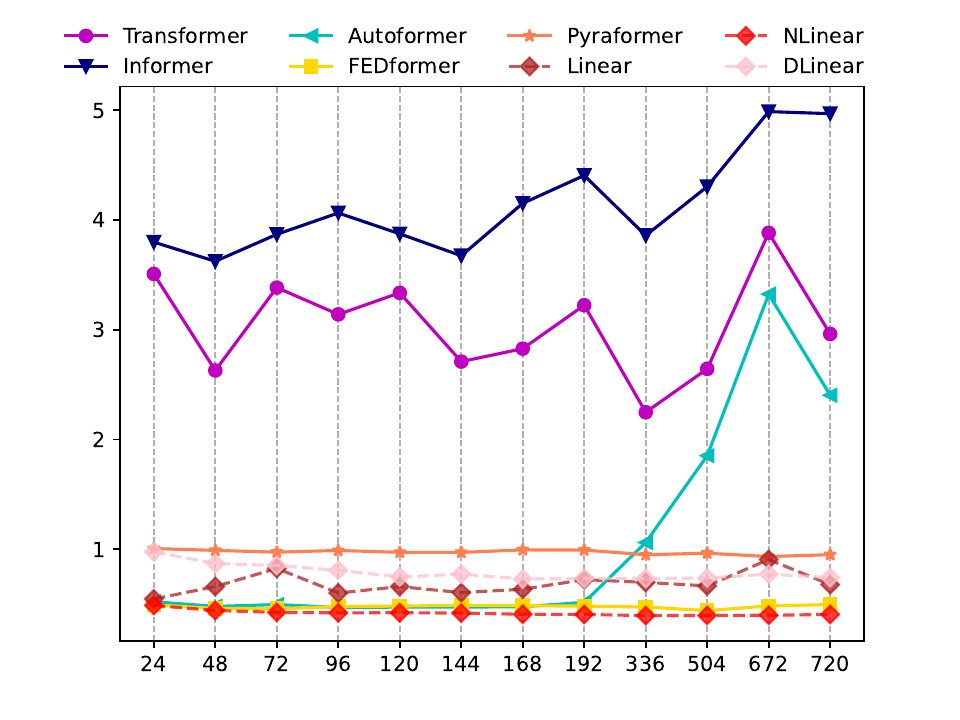}   
		\end{minipage}
	}
		\subfigure[\textbf{24} steps-\textbf{ETTm1}]
	{
		\begin{minipage}{4.1cm}
			\centering     
			\includegraphics[scale=0.3]{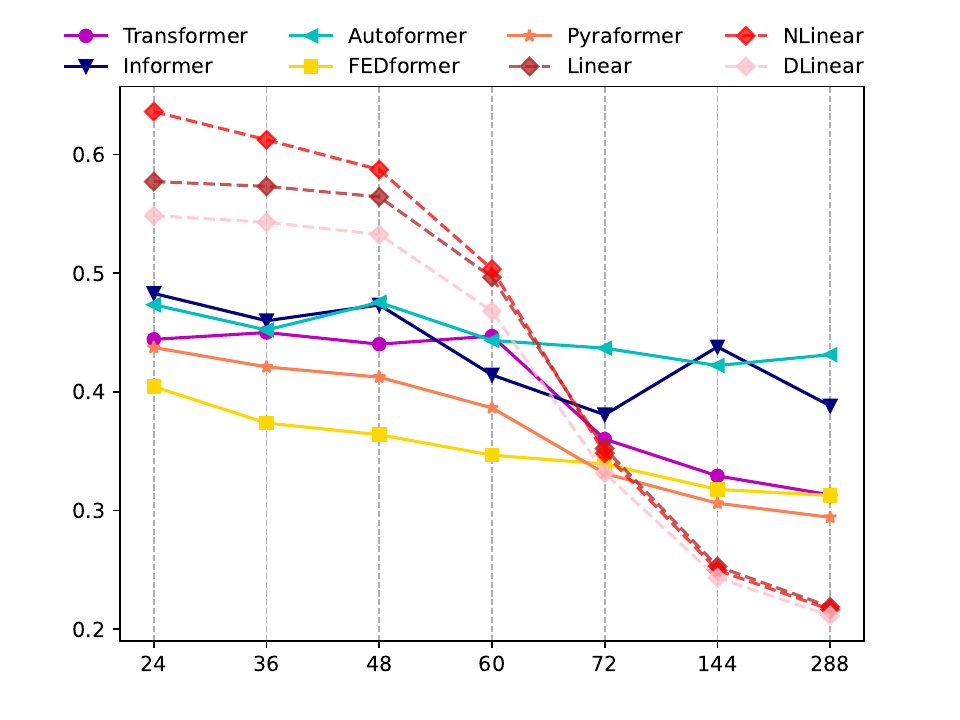}
		\end{minipage}
	}
	\subfigure[\textbf{576} steps-\textbf{ETTm1}]
	{
		\begin{minipage}{4.1cm}
			\centering     
			\includegraphics[scale=0.3]{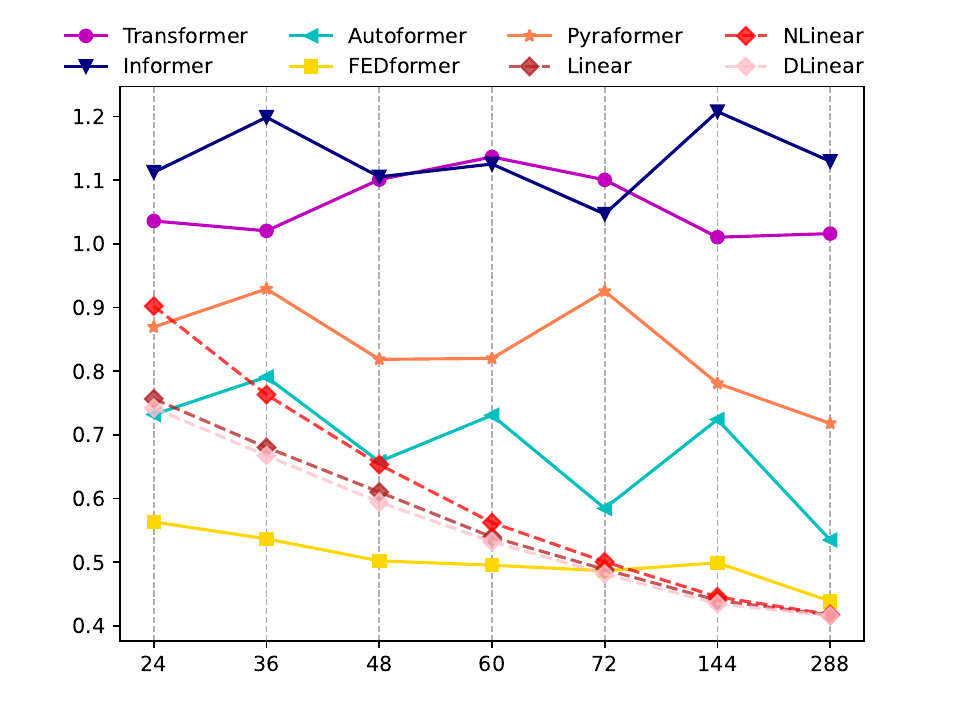} 
		\end{minipage}
	}
	\hspace{-2pt}
	\subfigure[\textbf{24} steps-\textbf{ETTm2}] 
	{
		\begin{minipage}{4.1cm}
			\centering         
			\includegraphics[scale=0.3]{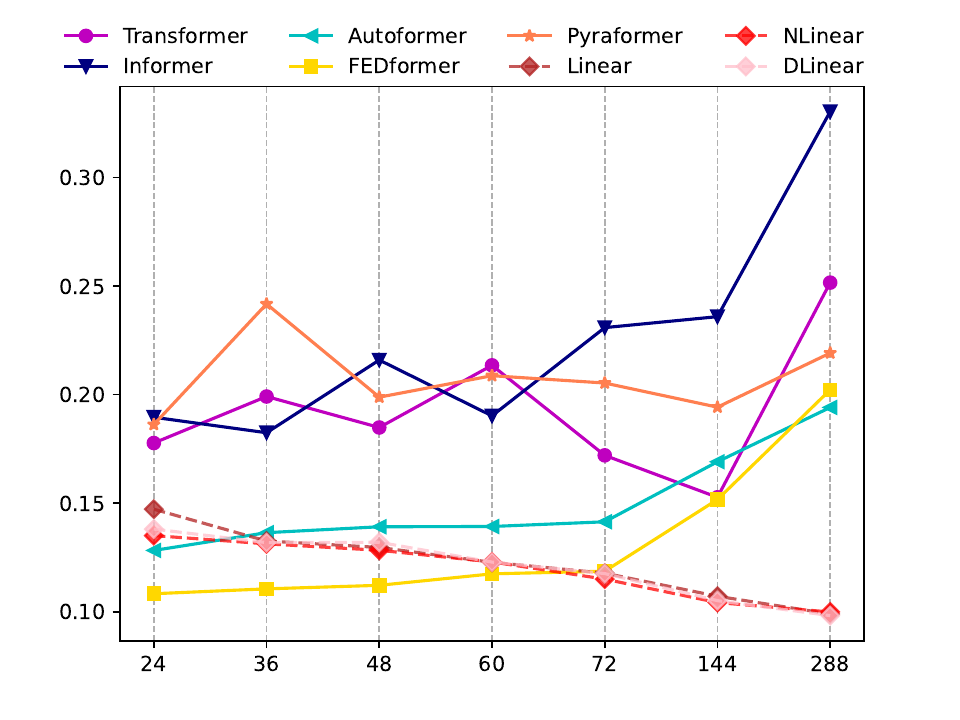} 
		\end{minipage}
	}
	\subfigure[\textbf{576} steps-\textbf{ETTm2}] 
	{
		\begin{minipage}{4.1cm}
			\centering      
			\includegraphics[scale=0.3]{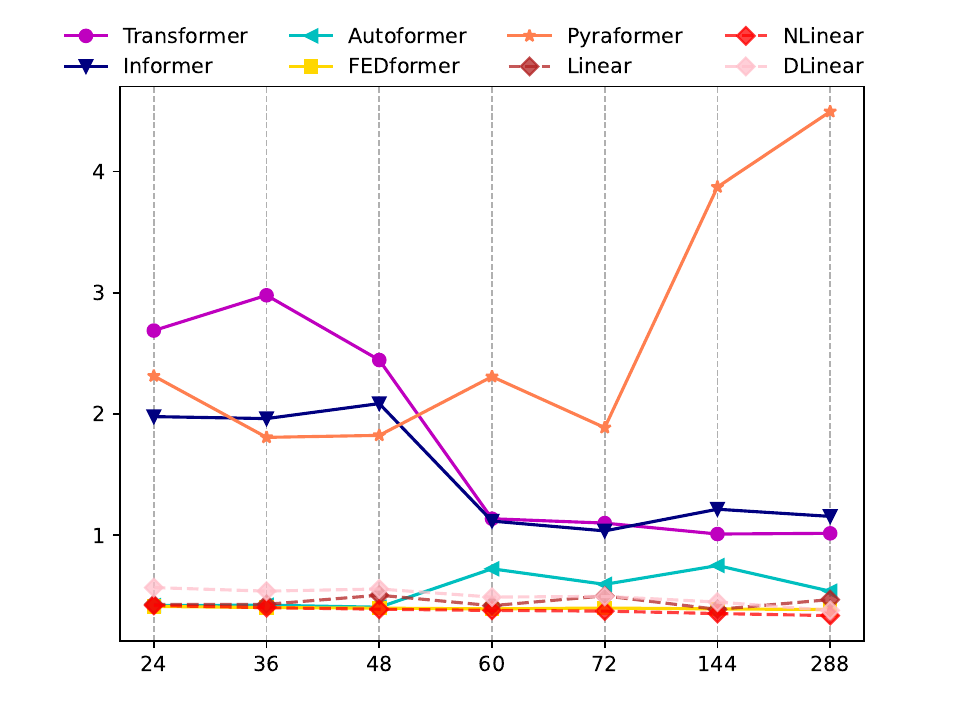}   
		\end{minipage}
	}
	\subfigure[\textbf{24} steps-\textbf{Weather}] 
	{
		\begin{minipage}{4.1cm}
			\centering         
			\includegraphics[scale=0.3]{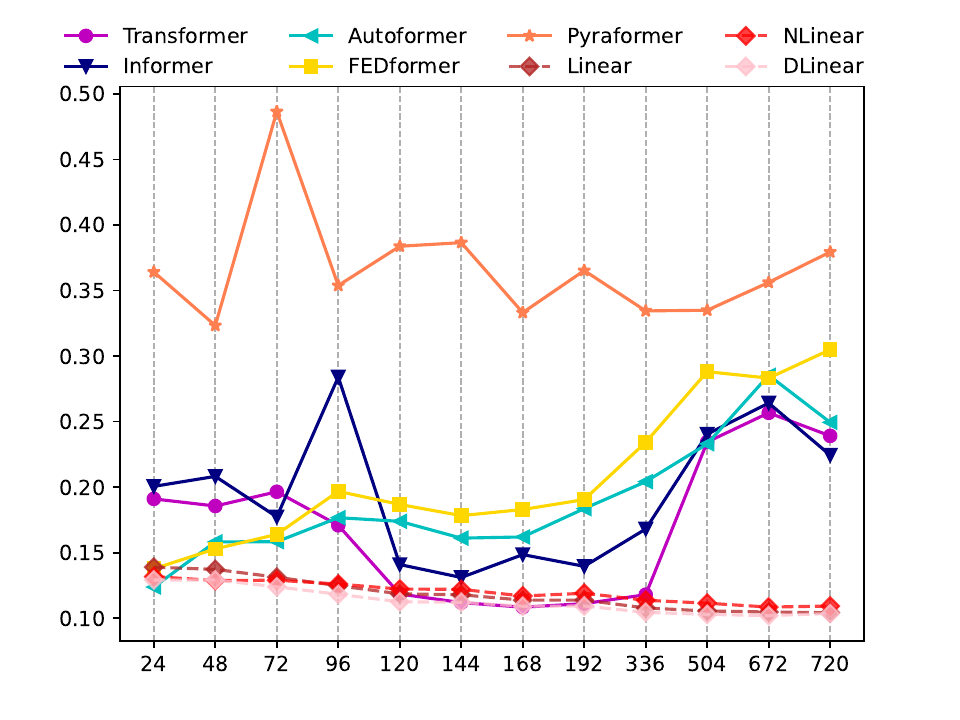} 
		\end{minipage}
	}
	\subfigure[\textbf{720} steps-\textbf{Weather}] 
	{
		\begin{minipage}{4.1cm}
			\centering      
			\includegraphics[scale=0.3]{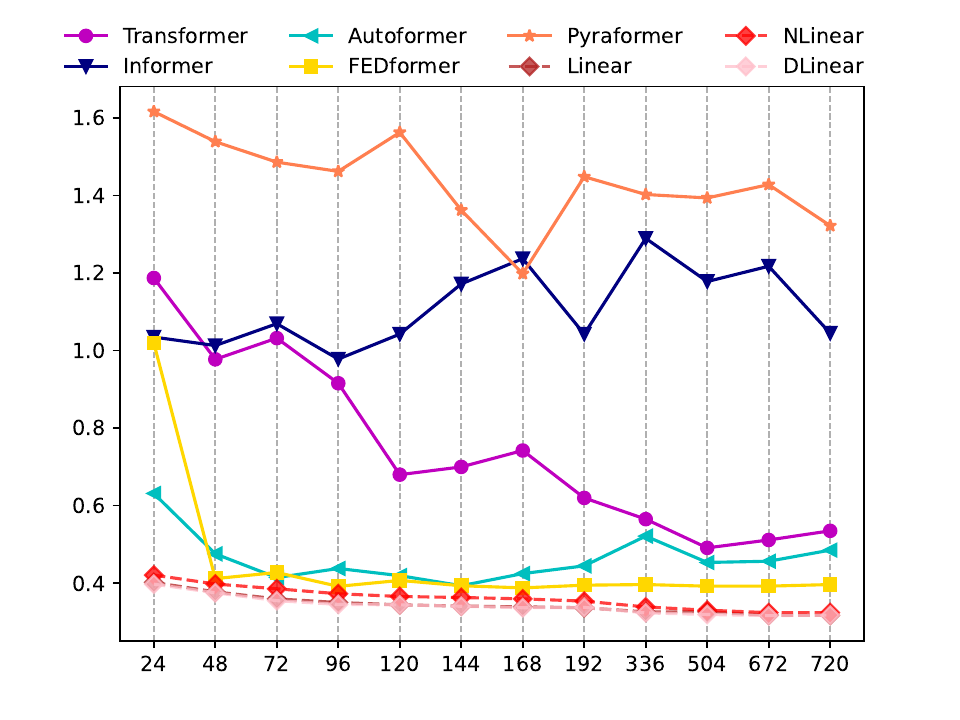}   
		\end{minipage}
	}
	\hspace{-2pt}
	\subfigure[\textbf{24} steps-\textbf{Traffic}]
	{
		\begin{minipage}{4.1cm}
			\centering     
			\includegraphics[scale=0.3]{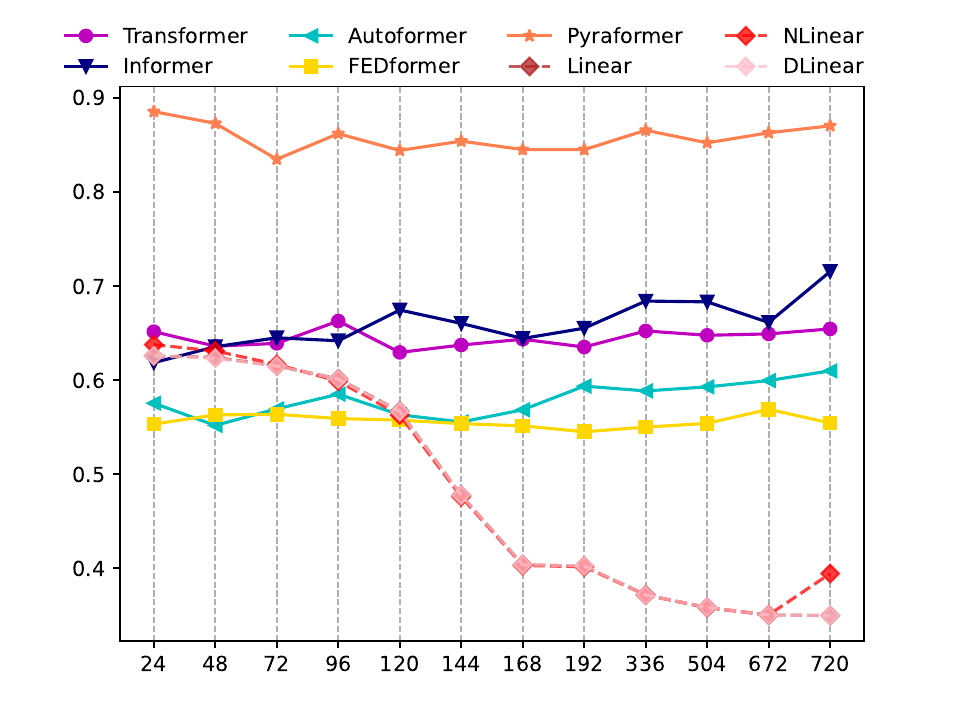}
		\end{minipage}
	}
	\subfigure[\textbf{720} steps-\textbf{Traffic}]
	{
		\begin{minipage}{4.1cm}
			\centering     
			\includegraphics[scale=0.3]{img/traffic_720.pdf} 
		\end{minipage}
	}
	\subfigure[\textbf{24} steps-\textbf{Exchange}] 
	{
		\begin{minipage}{4.1cm}
			\centering         
			\includegraphics[scale=0.3]{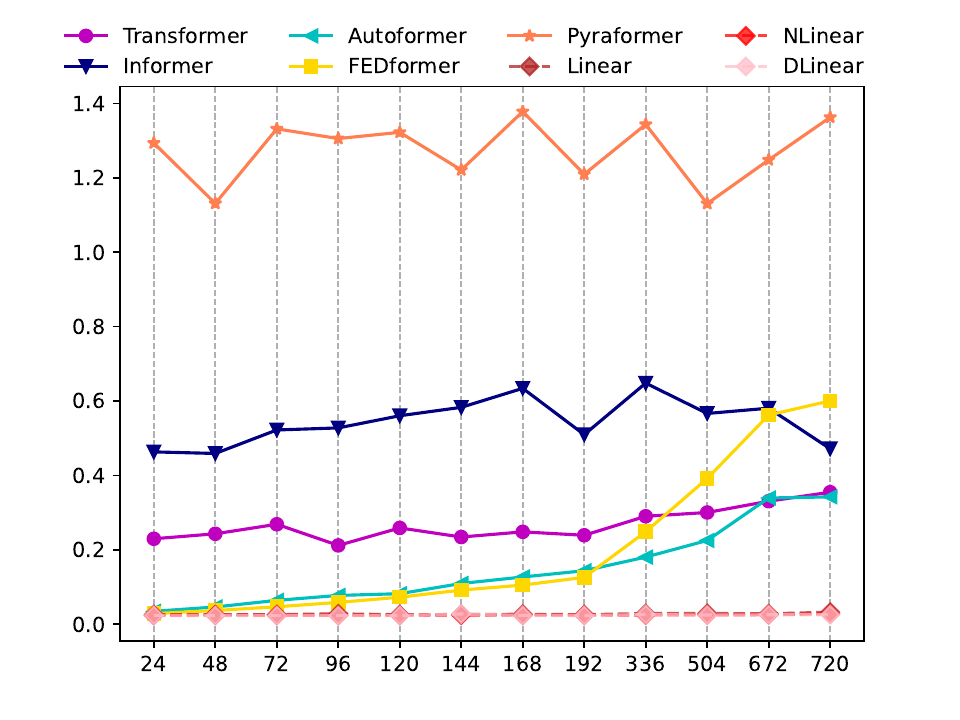} 
		\end{minipage}
	}
	\subfigure[\textbf{720} steps-\textbf{Exchange}] 
	{
		\begin{minipage}{4.1cm}
			\centering      
			\includegraphics[scale=0.3]{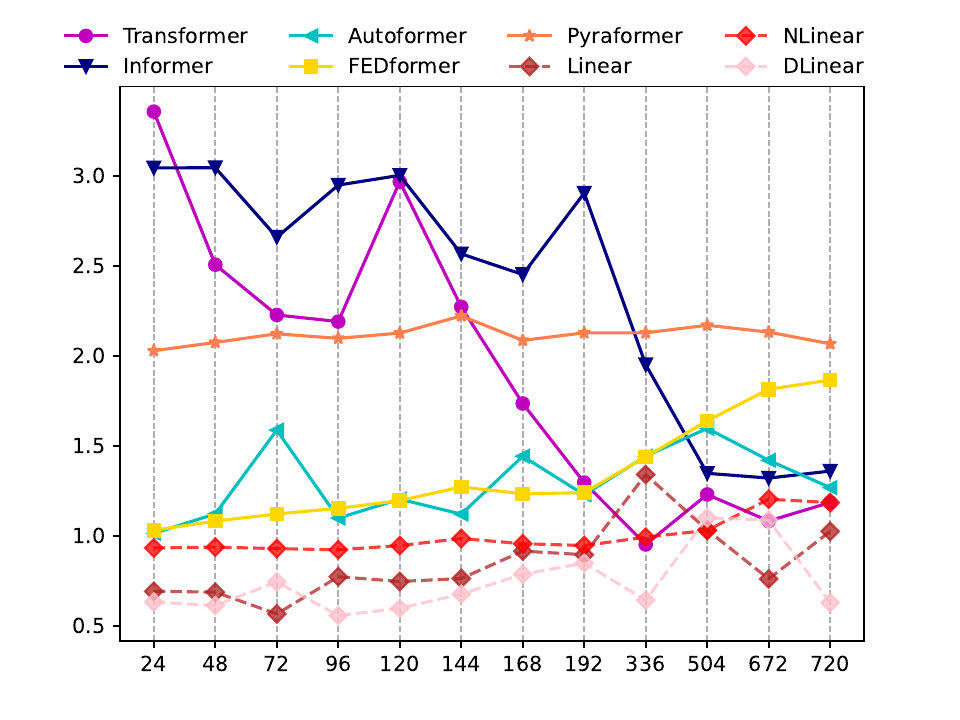}   
		\end{minipage}
	}
	\hspace{-2pt}
		\subfigure[\textbf{24} steps-\textbf{ILI}]
	{
		\begin{minipage}{4.1cm}
			\centering     
			\includegraphics[scale=0.3]{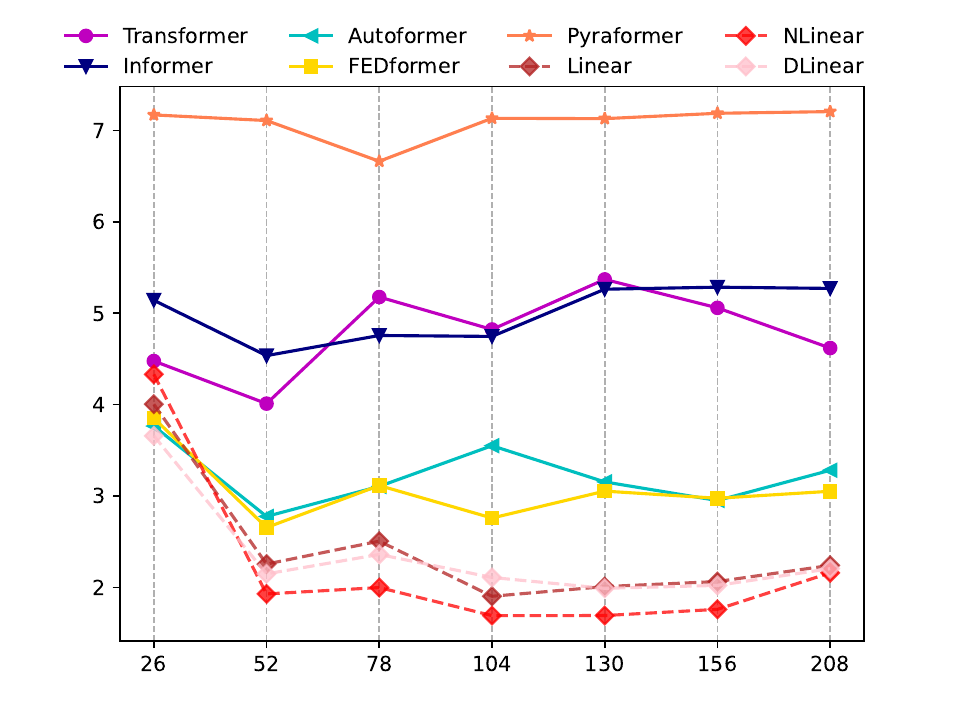}
		\end{minipage}
	}
	\subfigure[\textbf{60} steps-\textbf{ILI}]
	{
		\begin{minipage}{4.1cm}
			\centering     
			\includegraphics[scale=0.3]{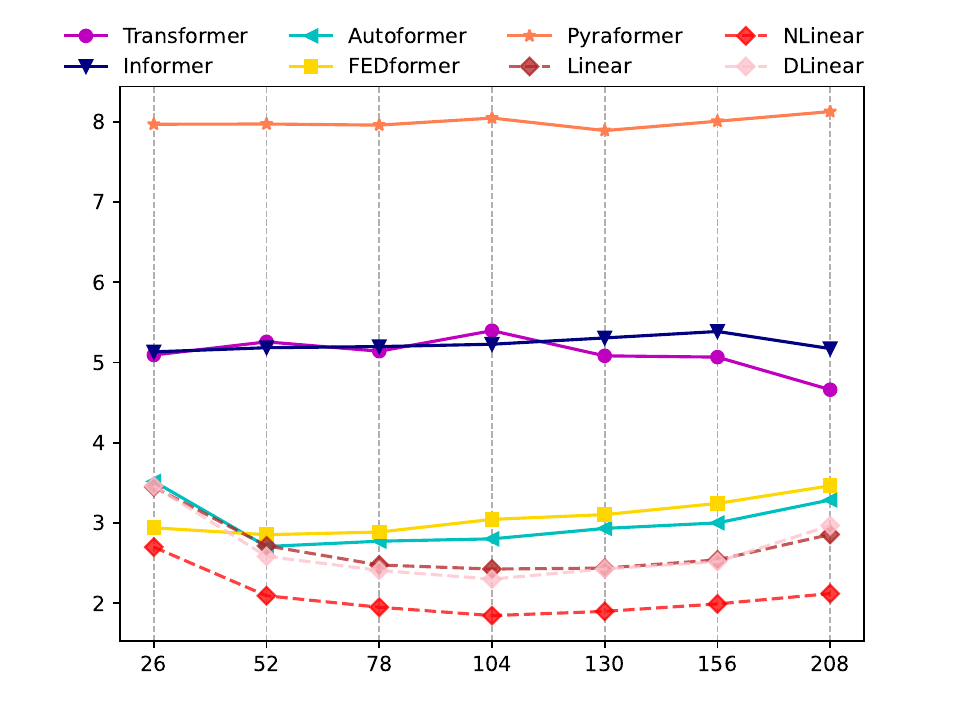} 
		\end{minipage}
	}
	\caption{The MSE results (Y-axis) of models with different look-back window sizes (X-axis) of the long-term forecasting (e.g., \textbf{720}-time steps) and the short-term forecasting (e.g., \textbf{24} time steps) on different benchmarks. } 
	\label{fig:supp_lookBackWindows} 
\end{figure*}

\subsection{Comparison under Different Look-back Windows}

In Figure~\ref{fig:supp_lookBackWindows}, we provide the MSE comparisons of five LTSF-Transformers with \modelname under different look-back window sizes to explore whether existing Transformers can extract temporal well from longer input sequences. 
For hourly granularity datasets (ETTh1, ETTh2, Traffic, and Electricity), the increasing look-back window sizes are \{24, 48, 72, 96, 120, 144, 168, 192, 336, 504, 672, 720\}, which represent \{1, 2, 3, 4, 5, 6, 7, 8, 14, 21, 28, 30\} days. The forecasting steps are \{24, 720\}, which mean \{1, 30\} days. For 5-minute granularity datasets (ETTm1 and ETTm2), we set the look-back window size as \{24, 36, 48, 60, 72, 144, 288\}, which represent \{2, 3, 4, 5, 6, 12, 24\} hours. For 10-minute granularity datasets (Weather), we set the look-back window size as \{24, 48, 72, 96, 120, 144, 168, 192, 336, 504, 672, 720\}, which mean \{4, 8, 12, 16, 20, 24, 28, 32, 56, 84, 112, 120\} hours. The forecasting steps are \{24, 720\} that are \{4, 120\} hours. For weekly granularity dataset (ILI), we set the look-back window size as \{26, 52, 78, 104, 130, 156, 208\}, which represent \{0.5, 1, 1.5, 2, 2.5, 3, 3.5, 4\} years. The corresponding forecasting steps are \{26, 208\}, meaning \{0.5, 4\} years.

As shown in Figure~\ref{fig:supp_lookBackWindows}, with increased look-back window sizes, the performance of \modelname is significantly boosted for most datasets (e.g., ETTm1 and Traffic), while this is not the case for Transformer-based TSF solutions. Most of their performance fluctuates or gets worse as the input lengths increase. To be specific, the results of Exchange-Rate do not show improved results with a long look-back window (from Figure~\ref{fig:supp_lookBackWindows}(m) and (n)), and we attribute it to the low information-to-noise ratio in such financial data.

\begin{figure*}[h]
\begin{center}
\includegraphics[width=1\textwidth]{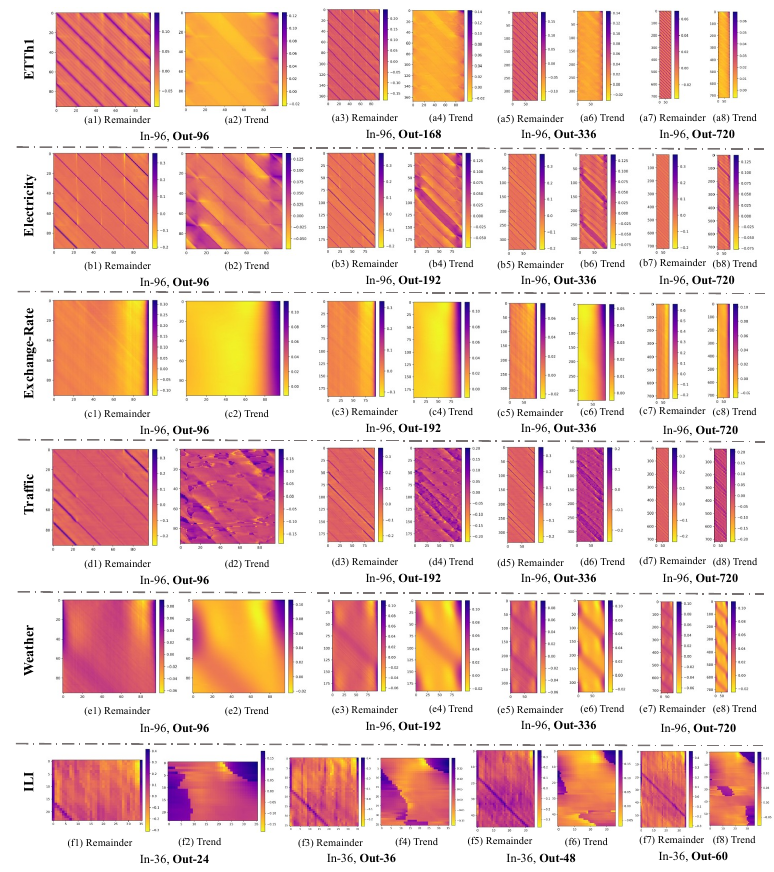}
\end{center}
\caption{Visualization of the weights(T*L) of \modelname on several benchmarks. Models are trained with a look-back window L (X-axis) and different forecasting time steps T (Y-axis). We show weights in the remainder and trend layer.}
\label{fig:viz}
\end{figure*}

\section{Ablation study on the \modelname}
\label{sec:supp_lookback}

\subsection{Motivation of NLinear}
If we normalize the test data by the mean and variance of train data, there could be a distribution shift in testing data, i.e, the mean value of testing data is not 0. If the model made a prediction that is out of the distribution of true value, a large error would occur. For example, there is a large error between the true value and the true value minus/add one. Therefore, in \emph{NLinear}, we use the subtraction and addition to shift the model prediction toward the distribution of true value. Then, large errors are avoided, and the model performances can be improved.
Figure~\ref{fig:distribution_shift} illustrates histograms of the trainset-test set distributions, where each bar represents the number of data points. Clear distribution shifts between training and testing data can be observed in ETTh1, ETTh2, and ILI. Accordingly, from Table~\ref{tab:uni-benchmarks-ett} and Table 2 in the main paper, we can observe that there are great improvements in the three datasets comparing the \emph{NLinear} to the \emph{Linear}, showing the effectiveness of the \emph{NLinear} in relieving distribution shifts. Moreover, for the datasets without obvious distribution shifts, like Electricity in Figure~\ref{fig:distribution_shift}(c), using the vanilla \emph{Linear} can be enough, demonstrating the similar performance with \emph{NLinear} and \emph{DLinear}.

\subsection{The Features of LTSF-Linear}

Although \modelname is simple, it has some compelling characteristics:

\begin{itemize}
\item \textbf{An $O(1)$ maximum signal traversing path length}: The shorter the path, the better the dependencies are captured~\cite{liu2021pyraformer}, making \modelname capable of capturing both short-range and long-range temporal relations.

\item \textbf{High-efficiency:} As \modelname is a linear model with two linear layers at most, it costs much lower memory and fewer parameters and has a faster inference speed than existing Transformers (see Table $8$ in main paper).

\item \textbf{Interpretability:} After training, we can visualize weights from the seasonality and trend branches to have some insights on the predicted values~\cite{dong2008granular}.

\item \textbf{Easy-to-use:} \modelname can be obtained easily without tuning model hyper-parameters.

\end{itemize}

\subsection{Interpretability of LTSF-Linear} 
Because \modelname is a set of linear models, the weights of linear layers can directly reveal how \modelname works. The weight visualization of \modelname can also reveal certain characteristics in the data used for forecasting.

Here we take DLinear as an example. Accordingly, we visualize the trend and remainder weights of all datasets with a fixed input length of 96 and four different forecasting horizons. To obtain a smooth weight with a clear pattern in visualization, we initialize the weights of the linear layers in DLinear as $1/L$ rather than random initialization. That is, we use the same weight for every forecasting time step in the look-back window at the start of training. 

\textbf{How the model works:} Figure~\ref{fig:viz}(c) visualize the weights of the trend and the remaining layers on the Exchange-Rate dataset. Due to the lack of periodicity and seasonality in financial data, it is hard to observe clear patterns, but the trend layer reveals greater weights of information closer to the outputs, representing their larger contributions to the predicted values.

\textbf{Periodicity of data:} For Traffic data, as shown in Figure~\ref{fig:viz}(d), the model gives high weights to the latest time step of the look-back window for the {0,23,47...719} forecasting steps. Among these forecasting time steps, the {0, 167, 335, 503, 671} time steps have higher weights. Note that 24 time steps are a day, and 168 time steps are a week. This indicates that Traffic has a daily periodicity and a weekly periodicity.

\bibliographystyle{ieee_fullname}

\end{document}